\documentclass[lettersize,journal]{IEEEtran}
\usepackage{amsmath,amsfonts}
\usepackage{algorithmic}
\usepackage{algorithm}
\usepackage{array}
\usepackage{textcomp}
\usepackage{stfloats}
\usepackage{url}
\usepackage{verbatim}
\usepackage{graphicx}
\usepackage{subfigure}
\usepackage{cite}
\usepackage[breaklinks,colorlinks,
linkcolor=red,
anchorcolor=blue,
citecolor=blue
]{hyperref}
\usepackage[utf8]{inputenc} 
\usepackage[T1]{fontenc}    
\usepackage{url}            
\usepackage{booktabs}       
\usepackage{nicefrac}       
\usepackage{microtype}      
\usepackage{longfbox}
\usepackage{amssymb}
\usepackage{amsmath}   
\usepackage{mathtools}
\usepackage{amsthm}
\usepackage{bbm}
\usepackage{wrapfig}
\usepackage{multirow}
\usepackage{makecell}
\usepackage{parnotes}
\usepackage{float}
\usepackage[makeroom]{cancel}
\usepackage{chngcntr}
\usepackage{apptools}
\usepackage{enumitem}
\usepackage{framed,enumitem} 
\usepackage[most]{tcolorbox}
\usepackage[algo2e,algoruled,boxed,lined,norelsize]{algorithm2e} 
\usepackage{threeparttable}
\usepackage{subcaption}
\usepackage{tablefootnote}

\graphicspath{{figures/}}

\DeclareMathOperator{\softplus}{\texttt{\textbf{SoftPlus}}}
\newcommand{\norm}[1]{\left\lVert#1\right\rVert}

\newcommand*{\tran}{^{\mkern-1.5mu\mathsf{T}}}

\newtheorem{lemma}{Lemma}

\begin{document}

\title{
\LARGE \bf
Complementarity-Free Multi-Contact Modeling and Optimization \\for Dexterous Manipulation}

\makeatletter
\g@addto@macro\@maketitle{
  \vspace{-10pt}
\setcounter{figure}{0}
  \begin{figure}[H]
  \setlength{\linewidth}{\textwidth}
  \setlength{\hsize}{\textwidth}
  \centering
  \includegraphics[width=1\linewidth]{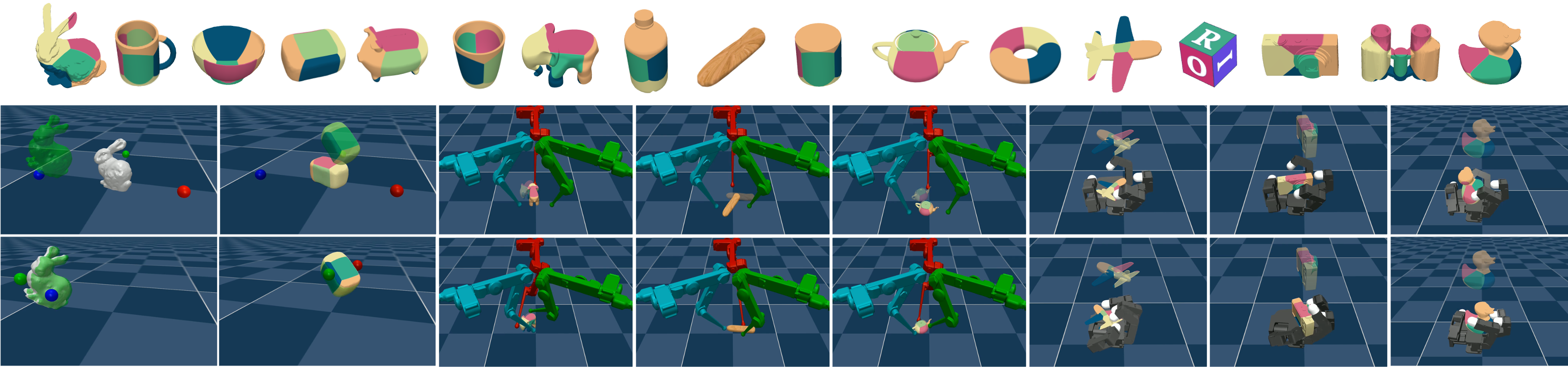}
  \caption{\small{We propose a complementarity-free multi-contact model that achieves state-of-the-art performance in planning and control across various challenging dexterous manipulation tasks, including fingertip in-air manipulation (cols. 1-2), TriFinger in-hand manipulation (cols. 3-5), and Allegro hand on-palm reorientation (cols. 6-8), all with diverse objects (first row). The second and third rows show the initial and final scenes of selected tasks, with the target object pose shown in transparency. Object diameters range from 50 [mm] to 150 [mm]. \href{https://youtu.be/NsL4hbSXvFg}{Video link}.
}}
  \label{fig.intro}
  \end{figure}
  \vspace{-30pt}
}
\makeatother

\author{Wanxin Jin \\
Arizona State University \\[3pt]
\href{https://youtu.be/NsL4hbSXvFg}{Video}
\,
\href{https://github.com/asu-iris/Complementarity-Free-Dexterous-Manipulation}{Code}
\,
\href{https://wanxinjin@gmail.com}{Email}
\\ 
}

\maketitle

\begin{abstract}
A significant barrier preventing model-based methods from achieving real-time and versatile dexterous robotic manipulation is the inherent complexity of multi-contact dynamics. Traditionally formulated as complementarity models, multi-contact dynamics introduces  non-smoothness and combinatorial complexity, complicating contact-rich planning and  optimization. In this paper, we circumvent these challenges by introducing a lightweight yet capable multi-contact model. Our new model, derived from the duality of optimization-based contact models, dispenses with the complementarity constructs entirely, providing computational advantages such as closed-form time stepping, differentiability, automatic satisfaction with Coulomb’s friction law, and minimal hyperparameter tuning. We demonstrate the model’s effectiveness and efficiency for planning and control in a range of challenging dexterous manipulation tasks, including fingertip 3D in-air  manipulation, TriFinger in-hand manipulation, and Allegro hand on-palm reorientation, all performed with diverse objects. Our method consistently achieves state-of-the-art results: (I) a 96.5\% average success rate across all objects and tasks, (II) high manipulation accuracy with an average reorientation error of $11^{\circ}$ and position error of $7.8 \text{mm}$, and (III) contact-implicit model predictive control running at 50-100 Hz for all objects and tasks. These results are achieved  with minimal hyperparameter tuning. 

\end{abstract}

\IEEEpeerreviewmaketitle


\section{Introduction}
Achieving real-time and versatile dexterity in robotic manipulation, which requires fast reasoning about frequent making and breaking contacts with various objects and physical environments, remains a significant challenge. Notable results have been delivered by reinforcement learning \cite{andrychowicz2020learning,rajeswaran2017learning,allshire2022transferring,chen2022system}, albeit at the expense of extensive environment data, lengthy training process, and limited object generalizability. 
 In contrast,  progress in model-based planning and control methods  been relatively slow, with existing methods typically limited to  simple  contact-less manipulation settings.  A primary challenge for model-based methods is the non-smooth and hybrid nature of contact-rich dynamics --- smooth motions are frequently interrupted by discrete contact events (i.e., making or breaking contacts) \cite{pang2023global,jin2024task}. This introduces computational challenges in both learning of contact dynamics  \cite{pfrommer2021contactnets} and combinatorics optimization  of contact modes \cite{cheng2022contact,pang2023global}.
Recent work has made some headway, including new  algorithms for learning hybrid dynamics \cite{pfrommer2021contactnets,jin2022learning}, differentiable contact modeling \cite{howell2022dojo,suh2022bundled,zhang2023adaptive},  multi-contact predictive control \cite{aydinoglu2022real,yang2024dynamic,kurtz2023inverse,le2024fast}, and contact-rich planning  \cite{cheng2022contact,pang2023global}. Despite these efforts, the computational challenges induced by non-smooth contact dynamics remain a significant barrier for model-based methods to achieve the real-time versatility needed for complex dexterous manipulation involving diverse object geometries and robot morphologies.

This paper seeks to advance the state-of-the-art in model-based dexterous manipulation by addressing a critical question: \emph{Can we circumvent the computational challenges induced by the  non-smooth multi-contact dynamics in the early stage of contact modeling?}  Unlike existing efforts \cite{aydinoglu2022real,yang2024dynamic,kurtz2023inverse,le2024fast,cheng2022contact,pang2023global}, we do NOT pursue abstract learning architectures \cite{nagabandi2020deep,jin2024task}, propose artificial  smoothing strategies for existing contact dynamics \cite{howell2022dojo,suh2022bundled,zhang2023adaptive},
nor develop  planning algorithms that explicitly tackle the hybrid decision space. Instead, our contribution is \emph{a new multi-contact model that overcomes the non-smooth and hybrid computational complexities via a physics modeling perspective}, enabling significant speed and performance improvement in model-based dexterous manipulation.

Specifically, unlike traditional complementarity-based formulations \cite{coumans2020,todorov2012mujoco,drake,lee2018dart,tasora2016chrono,howell2022dojo} for modeling rigid-body contacts,  our proposed method \emph{modifies  the complementarity constructs  in the dual space of the optimization-based contact models \cite{anitescu2006optimization,todorov2012mujoco} into explicit forms}. This results in a new complementarity-free multi-contact model that has several computational advantages.  (I) \emph{Closed-form and differentiable time-stepping}: the next system state is a closed-form differentiable function of the current state and input, thus avoiding solving complementarity problems \cite{stewart2000implicit},  optimization \cite{anitescu2006optimization}, or residual equations \cite{howell2022dojo} for contact constraint resolution.  (II) \emph{Automatic satisfaction with Coulomb’s friction law in a single term}: the new model resolves the normal   and frictional contact forces using a \emph{single} term that automatically respects the Coulomb friction cone. This is  unlike the contact models \cite{geilinger2020add,liu2024quasisim,tassa2012synthesis,chatzinikolaidis2020contact} that handle  normal and frictional components independently,  leading to additional hyperparameters. (III) \emph{Fewer hyperparameters}: the proposed model has fewer parameters, making it easy to tune, and it also supports model auto-tuning using any learning framework.

The goal of the new contact model is not to compete with existing  high-fidelity physics  simulators   \cite{coumans2020,todorov2012mujoco,drake,lee2018dart,tasora2016chrono,howell2022dojo}, but rather to offer a lightweight, smooth,  yet capable surrogate that addresses the inherent computational challenges associated with existing complementarity-based models, particularly when used for contact-rich optimization and control. We integrate this model into model predictive control (MPC) framework for contact-implicit online planning and control. With minimal hyperparameter tuning, we achieve state-of-the-art performance across challenging dexterous manipulation tasks, including 3D in-air fingertip manipulation, TriFinger in-hand manipulation, and Allegro hand on-palm reorientation, all with diverse objects, as shown in Fig. \ref{fig.intro}. Our method  sets a new benchmark for model-based contact-rich dexterous manipulation:
\begin{itemize}[leftmargin=*]
    \item Highly versatile  dexterity:  96.5\% average success rate across all objects and  environments.
    \item High-accuracy dexterity:  average object reorientation error of $11^{\circ}$ and position error of $7.8 \text{mm}$.
    \item Real-time dexterity: contact-implicit MPC running at  50-100 Hz for all  objects and  robots.
\end{itemize}

\section{Related Works}

\subsection{Rigid Body Multi-contact Models}
\subsubsection{Nonconvex Complementarity Contact Models}
Rigid body contact dynamics is traditionally formulated using complementarity models \cite{stewardtrinklePolyhedron,stewart2000implicit,stewart1996implicit}: it enforces no interpenetration and no contact force at a distance. The Coulomb friction law, which governs sticking and sliding contacts \cite{stewart2000implicit,stewart1996implicit}, can also be expressed as complementarity constraints via the maximum dissipation principle \cite{stewart1998convergence}, leading to a nonlinear complementarity problem (NCP). Since the NCPs cannot be interpreted as the KKT conditions of a convex program, they are challenging to solve. To simplify computation, Coulomb friction cones are often approximated by polyhedral cones \cite{stewart1996implicit,todorov2012mujoco}, converting the NCP into a linear complementarity problem (LCP) \cite{stewart1996implicit}, for which mature solvers exist \cite{dirkse1995path}.

\smallskip

\subsubsection{Cone Complementarity Contact Models} In \cite{anitescu2006optimization}, Anitescu  proposed relaxing  the NCP formulation, by constraining  the contact velocity within a dual friction cone. Then,  the NCP becomes cone complementarity problem (CCP), which attains computational benefits such as fast convergence and solution    guarantees  \cite{anitescu2010iterative,tasora2011matrix}. A side effect of  CCP is that it allows for small normal motion at the contact even when bodies should remain in contact. This creates a  “boundary layer” whose size is proportional to time step and tangential velocity \cite{anitescu2010iterative}. 

\smallskip

\subsubsection{Optimization-based Contact Models}
The CCP described above corresponds to the KKT optimality conditions of a convex optimization problem with second-order cone constraints \cite{anitescu2006optimization}. This allows for the formulation of a primal convex optimization (with velocities as decision variables) or a dual optimization (with contact forces as decision variables). The primal approach has been used in work \cite{pang2023global,pang2021convex}, while the dual formulation is used in the MuJoCo simulator \cite{todorov2012mujoco}, albeit with a regularization term for model invertibility. The dual objective function can be interpreted as minimizing kinetic energy in  contact frames.

\smallskip

\subsubsection{Differentiable Contact Models} Several alternative models have been proposed to approximate complementarity-based models for differentiability. \cite{geilinger2020add} introduced penalty functions to model contact normal and frictional forces.  \cite{howell2022dojo} proposes relaxing complementarity constraint, which in optimization-based model case, is equivalent to solving an unconstrained optimization with  constraints penalized via log-barrier functions \cite{jin2021safe}. 
\cite{todorov2010implicit} proposed implicit complementarity, converting all constraints to  unconstrained optimization with intermediate variables. \cite{yamane2006stable,liu2024quasisim} developed penalty-based contact models. A  shared feature   is that those methods ultimately need to solve a residual equation for contact constraint resolution, and differentiability is obtained via implicit function theorem \cite{rudin1964principles}.

\smallskip

\subsubsection{Why  is our model new?} Two features distinguish the proposed model from existing ones. First, closed-form contact constraint resolution: our model builds on  optimization-based contact dynamics \cite{anitescu2006optimization,pang2021convex}, but instead of solving the primal \cite{anitescu2006optimization,pang2021convex} or dual programs \cite{todorov2012mujoco}, we approximate  the complementarity constraints  in the dual space with an explicit form (by contact decoupling). This approximation leads to a \emph{closed-form} contact constraint resolution, eliminating the need to solve optimization \cite{anitescu2006optimization}, residual \cite{howell2022dojo}, or complementarity problems \cite{stewart2000implicit} at each time stepping.

Second, unified handling of contact normal and friction: while previous work \cite{tassa2012synthesis,chatzinikolaidis2020contact,yamane2006stable, liu2024quasisim} has also explored  closed-form contact dynamics models, our novelty, however, is that \emph{our closed-form model is approximated in the dual friction cone}. This allows for  the \emph{unified} treatment of contact normal forces and frictional forces within a single term, ensuring automatic sanctification with Coulomb’s friction law. In contrast, existing  models \cite{tassa2012synthesis,chatzinikolaidis2020contact,yamane2006stable, liu2024quasisim}  treat the  contact normal and enforce  Coulomb’s friction separately, which typically leads to additional hyperparameters and potential nonphysical artifacts.

\vspace{-5pt}
\subsection{Planning and Control with Contact Dynamics}
Planning and control for multi-contact systems are challenging because algorithms must determine when and where to make or break contacts, with complexity scaling exponentially with potential contact locations and planning horizons.  Traditional  methods \cite{sleiman2021unified,mastalli2020crocoddyl}  predefine contact sequences, which could be possible for tasks like legged locomotion \cite{winkler2018gait}.  Modern approaches focus  on contact-implicit planning \cite{posa2014direct}, solving for  both contact location and sequencing.  Two main strategies are proposed. The first is to smooth  contact transition boundaries. In \cite{le2024fast,posa2014direct,pang2023global,mordatch2012discovery}  complementarity constraints are relaxed. The other is to maintain the hybrid structures and cast the planning as mixed integer program, as done in  \cite{aydinoglu2022real,aydinoglu2023consensus,pang2023global,aceituno2020global,yang2024dynamic,huang2024adaptive}.

This paper develops the  contact-implicit planning and control  based on the proposed complementarity-free contact model. The resulting contact-implicit optimization can be readily solved using standard optimization techniques \cite{li2004iterative} or MPC tools \cite{Andersson2019}, requiring no additional smoothing constructs or  optimization programming effort.
With its closed form and smooth nature, our model significantly improves the speed of  online planning; e.g., our contact-implicit MPC controller runs up to  50-100 Hz in the Allegro Hand dexterous reorientation tasks.

\vspace{-5pt}
\subsection{Reinforcement Learning for Dexterous Manipulation}
Reinforcement learning (RL) has shown impressive results in dexterous manipulation \cite{andrychowicz2020learning,rajeswaran2017learning,allshire2022transferring,chen2022system}. For instance, \cite{andrychowicz2020learning,allshire2022transferring} employ model-free RL for in-hand object reorientation; \cite{qi2023hand} introduces an adaptive framework for reorienting various objects. However, these methods require millions to billions of environment samples. Model-based RL \cite{jin2024task,nagabandi2020deep} offers better efficiency, but unstructured deep models can struggle with multimodality \cite{nagabandi2020deep,pfrommer2021contactnets}. The prior work \cite{jin2024task} shows incorporating hybrid constructs into models significantly improves efficiency, enabling dexterous manipulation with only thousands of environment samples.

RL has set the state-of-the-art in challenging dexterous manipulation tasks, e.g., TriFinger \cite{jin2024task,allshire2022transferring} and Allegro hand manipulation \cite{qi2023hand}, where model-based methods often struggle. Our proposed method aims to bridge this gap and even surpass state-of-the-art RL in success rate and manipulation accuracy.

\section{Preliminary and Problem Statement}
We consider a manipulation system, comprised of an actuated robot,  unactuated object and environment (e.g., ground). We define the following notations  used in the rest of the paper.
\smallskip

\noindent
\begin{tabular}{ll}
\small
$\boldsymbol{q}_o\in\mathbb{R}^{n_{q_o}}$ & object position \\
$\boldsymbol{v}_o\in\mathbb{R}^{n_o}$ & object velocity \\
$\boldsymbol{\tau}_o\in\mathbb{R}^{n_o}$ &   non-contact force applied to object \\

$\boldsymbol{q}_r\in\mathbb{R}^{n_r}$ &robot position (generalized coordinate) \\
$\boldsymbol{v}_r\in\mathbb{R}^{n_r}$ & robot velocity (generalized coordinate)  \\
$\boldsymbol{\tau}_r\in\mathbb{R}^{n_r}$ &   non-contact force  to robot \\
$\boldsymbol{u} \in\mathbb{R}^{n_u}$ &  control input applied to robot\\

$\boldsymbol{q} {=}(\boldsymbol{q}_o, \boldsymbol{q}_r )$ & system position \\
$\boldsymbol{v} {=}  (\boldsymbol{v}_o, \boldsymbol{v}_r )$ & system velocity\\

$\boldsymbol{\lambda}{=}(\lambda^{n}, \boldsymbol{\lambda}^d)$ &contact  impulse/force (normal, friction) \\

$\boldsymbol{J}_r $  &   contact Jacobian of the robot  \\
$\boldsymbol{J}_o  $  &  contact Jacobian of the object\\
$  \boldsymbol{J}  =[ \boldsymbol{J}_o, \boldsymbol{J}_r] $ & {contact Jacobian  of the system}\\
$ \boldsymbol{J}  {=}[\boldsymbol{J}^n; \boldsymbol{J}^d]$ & {$\boldsymbol{J}$ reformatted in normal \& tangential}\\
$ \boldsymbol{J}^d  {=}[\boldsymbol{J}^d_1;..; \boldsymbol{J}^d_{n_d} ]$ & {tangential Jacobian for polyhedral cone}

\end{tabular}

\subsection{Optimization-based Quasi-Dynamic Contact Model}
For simplicity, we model a manipulation system using the quasi-dynamic formulation \cite{mason2001mechanics,cheng2022contact,aceituno2020global,pang2023global}, which primarily captures the positional displacement of a contact-rich system in relation to contact interactions and  inputs, while ignoring the  inertial and Coriolis forces that are less significant under less dynamic motion. Quasi-dynamic models benefit from simplicity and suffice for a wide variety of manipulation tasks \cite{pang2023global,cheng2022contact,aceituno2020global}. It should be noted our following model can be readily extended to full  dynamic model, which will be shown in Section \ref{sec.sim_dyn_setting}.

Formally, consider a manipulation system with $n_c$ potential contacts, which could happen between the robot and object or/and   between object and the environment. The  time-stepping  equation of the quasi-dynamic model is
\begin{equation}\label{equ.quasi_dyn}
\begin{gathered}
    \epsilon\boldsymbol{M}_o\boldsymbol{v}_o = 
    h\boldsymbol{\boldsymbol{\tau}}_o + \sum\nolimits_{i=1}^{n_c}{\boldsymbol{J}}\tran_{o,i}\boldsymbol{\lambda}_i,
    \\
    h\boldsymbol{K}_r(h\boldsymbol{v}_r - \boldsymbol{u}) = h\boldsymbol{\boldsymbol{\tau}}_r + \sum\nolimits_{i=1}^{n_c}{\boldsymbol{J}}\tran_{r,i}\boldsymbol{\lambda}_i.
\end{gathered}
\end{equation}
Here,  $h$ is the time step. The first equation is the motion of the object with $\epsilon\boldsymbol{M}_o \in \mathbb{R}^{n_o \times n_o}$ the regularized mass matrix for the object  ($\epsilon>0$ is the regularization parameter)\cite{pang2022TRO}. The second equation is the motion of the  robot, where we follow \cite{pang2021convex} and consider the  robot is in  impedance control \cite{hogan1984impedance} and thus can be viewed as as a "spring" with  stiffness   $\boldsymbol{K}_r \in \mathbb{R}^{n_r \times n_r}$; the input 
$\boldsymbol{u}\in \mathbb{R}^{n_r}$ to the robot is the \emph{desired joint displacement}. $\boldsymbol{\lambda}_i=(\lambda^n_{i}, \boldsymbol{\lambda}^d_{i})\in\mathbb{R}^3$ is the $i$-th contact impulse, with contact normal component $\lambda_{i}^n$ and frictional component $\boldsymbol{\lambda}_{i}^d$. 

Next, we add   contact constraints between the contact impulse and system motion using the cone complementarity formulation proposed by  Anitescu in \cite{anitescu2006optimization,anitescu2010iterative}. Specifically,  the cone complementarity constraint at  contact $i$ writes
\begin{equation}\label{equ.cone_comple}
    \\ \mathcal{K}_i\ni \boldsymbol{\lambda}_i \perp \boldsymbol{J}_i\boldsymbol{v}+\frac{1}{h}\begin{bmatrix}
        {\phi_i}\\
        0\\
        0
    \end{bmatrix}\in \mathcal{K}_i^*, \quad 
    \forall i=\{1,...,n_c\},
\end{equation}
where $\phi_i$ is the normal distance at contact $i$; $\mathcal{K}_i$ is the Coulomb frictional cone, defined as $        \mathcal{K}_i=\left\{
        \boldsymbol{\lambda}_i\in \mathbb{R}^3\,\,|\,\,
        \mu_i\lambda_{i}^n\geq \norm{\boldsymbol{\lambda}_{i}^d}
        \right\}$;
 $\mathcal{K}_i^*$ is the dual cone to   $\mathcal{K}_i$, and the right side of (\ref{equ.cone_comple}) says that at contact $i$, the relaxed contact velocity has to lie in  $\mathcal{K}_i^*$, i.e., 
\begin{equation}\label{equ.second-order-cone}
    \begin{aligned}
        \boldsymbol{v}\in\bigg\{
         \boldsymbol{v}\,\,|\,\,
        \boldsymbol{J}^n_i\boldsymbol{v}+\frac{\phi_i}{h}\geq \mu_i\norm{\boldsymbol{J}^d_i\boldsymbol{v}}
        \bigg\}
    \end{aligned},
\end{equation}
with 
 $\mu_i$  the friction coefficient for contact $i$. 

In \cite{anitescu2006optimization}, Anitescu showed  (\ref{equ.quasi_dyn}) and (\ref{equ.cone_comple}) are the KKT optimality conditions for the following primal optimization \cite{anitescu2006optimization}:
\begin{equation}\label{equ.qs_model}
\begin{aligned}
    \min_{\boldsymbol{v}} \quad &\frac{1}{2}h^2\boldsymbol{v}\tran\boldsymbol{Q}\boldsymbol{v}-{h} \boldsymbol{v}\tran\boldsymbol{b}(\boldsymbol{u})\\
    \text{subject to} \quad & 
\boldsymbol{J}_i\boldsymbol{v}+\frac{1}{h}\begin{bmatrix}
        {\phi_i}\\
        0\\
        0
    \end{bmatrix}\in \mathcal{K}_i^*, \quad   i \in\{1...n_c\},
\end{aligned}
\end{equation}
where $\boldsymbol{Q} \in \mathbb{R}^{(n_o+n_r) \times (n_o+n_r)}$ and $\boldsymbol{b}(\boldsymbol{u}) \in \mathbb{R}^{n_o+n_r}$ are 
\begin{equation}\label{equ.q_matrix}
\boldsymbol{Q}:=\begin{bmatrix}
    \epsilon \boldsymbol{M}_o/h^2 & \boldsymbol{0}\\
    \boldsymbol{0}& \boldsymbol{K}_r
\end{bmatrix},
\quad 
    \boldsymbol{b}(\boldsymbol{u}):=\begin{bmatrix}
        \boldsymbol{\tau}_o\\
        \boldsymbol{K}_r \boldsymbol{u}+\boldsymbol{\tau}_r
    \end{bmatrix},
\end{equation}
 respectively. (\ref{equ.qs_model}) is a second-order cone program (SOCP). To facilitate solving the SOCP, one can approximate it with a quadratic program (QP) by linearizing the second-order dual cone constraint using its polyhedral approximation \cite{stewardtrinklePolyhedron}.   Specifically, at  contact $i$, one can use a symmetric set of $n_d$ unit directional vectors $\left\{\boldsymbol{d}_{i,1}, \boldsymbol{d}_{i,2}, ...,\boldsymbol{d}_{i,n_d}\right\}$ to span the contact tangential plane  \cite{stewart2000implicit}, yielding the   linearization of (\ref{equ.second-order-cone}):
\begin{equation}\label{equ.linearized_dual_cone}
    \boldsymbol{J}^n_i\boldsymbol{v}+\frac{\phi_i}{h}\geq \mu_i\boldsymbol{J}^d_{i,j}\boldsymbol{v}, \quad  \forall j \in\{1...n_d\}
\end{equation}
with $\boldsymbol{J}^d_{i,j}$ being the system Jacobian to the unit directional vector $\boldsymbol{d}_j$ in the tangential plane of  contact $i$. Hence, the SOCP in  (\ref{equ.qs_model}) can be simplified as the following QP
\begin{equation}\label{equ.primal_prob}
\begin{aligned}
    \min_{\boldsymbol{v}} \quad &\frac{1}{2}h^2\boldsymbol{v}\tran\boldsymbol{Q}\boldsymbol{v}-{h} \boldsymbol{v}\tran\boldsymbol{b}(\boldsymbol{u})\\
    \text{subject to} \quad & 
(\boldsymbol{J}_{i}^n-\mu_i\boldsymbol{J}_{i,j}^{d})\boldsymbol{v}+\frac{\phi_i}{h}\geq 0, \\
& i \in\{1...n_c\},\,\,\,j \in\{1...n_d\}.
\end{aligned}
\end{equation}
In  time-stepping  prediction,   Jacobians $\boldsymbol{J}^n_{i}$ and $\boldsymbol{J}^d_{i,j}$ 
 are calculated from a collision detection routine \cite{pan2012fcl} at the  system's current position $\boldsymbol{q}$. The solution $\boldsymbol{v}^{+}$ to (\ref{equ.primal_prob}) will be used to integrate from $\boldsymbol{q}$ to the next position $\boldsymbol{q}^{+}$, and we simply write it as $\boldsymbol{q}^{+}=\boldsymbol{q}\oplus h\boldsymbol{v}^+$ ($\oplus$ can involve quaternion integration).

\subsection{Model Predictive Control}
The generic formulation of model predictive control is
\begin{equation}\label{equ.generic_mpc}
\begin{aligned}
\min_{\boldsymbol{u}_{0:T-1}\in[\mathbf{u}_{\text{lb}}, \mathbf{u}_{\text{ub}}]} \quad 
&  
\sum\nolimits_{t=0}^{T-1} \textit{c}\left ( \boldsymbol{q}_t, \boldsymbol{u}_t \right ) +V\left (\boldsymbol{q}_T\right) 
\\ \text{subject to}\quad &
\boldsymbol{q}_{t+1} = \boldsymbol{f} \left ( \boldsymbol{q}_{t}, \boldsymbol{u}_{t} \right ), \ t = 0,...,T{-}1,\quad
\\ &\text{given} \ \boldsymbol{q}_0.
\end{aligned}
\end{equation}
where  model  $\boldsymbol{f}$ predicts the next system state. With $\boldsymbol{q}_0$,  (\ref{equ.generic_mpc}) searches for the optimal input sequence ($\mathbf{u}_{\text{lb}}$ and $\mathbf{u}_{\text{lb}}$ are control bounds), by minimizing the path  $\textit{c}(\cdot)$ and final cost $V(\cdot)$.

\smallskip 
In a manipulation system,  the MPC policy is implemented  in a receding horizon fashion, by repeatedly solving (\ref{equ.generic_mpc}) at the real system state $\boldsymbol{q}_k^{\text{real}}$ encountered at the policy rollout step $k$ and only applying the first optimal input to the real system. Specifically, at the encountered  system state  $\boldsymbol{q}_k^{\text{real}}$, the MPC policy sets $\boldsymbol{q}_0=\boldsymbol{q}_k^{\text{real}}$ and solves (\ref{equ.generic_mpc}). Only the first optimal input $\boldsymbol{u}_0^*(\boldsymbol{q}_k^{\text{real}})$ is applied to the real system,  evolving the  system state to the next $\boldsymbol{q}^{\text{real}}_{k+1}$. This  implementation creates a closed-loop control effect on the real system, i.e., feedback from system state $\boldsymbol{q}^{\text{real}}_{k}$ to control input $\boldsymbol{u}^*_0(\boldsymbol{q}^{\text{real}}_{k})$.

\subsection{Problem: Contact-implicit MPC for Dexterous Manipulation}

We are interested in  real-time, contact-implicit MPC  for dexterous manipulation. Directly using the  QP-based contact model  (\ref{equ.primal_prob})  in MPC (\ref{equ.generic_mpc}) leads to a nested optimization, which is difficult to solve due to the non-smooth behavior  of the contact model  (\ref{equ.primal_prob}), i.e., the prediction ``jumps" at the transitions between separate, sliding, and sticking contact modes. 
The goal of the paper is to develop a new  surrogate multi-contact model $\boldsymbol{f}$ in (\ref{equ.generic_mpc}) to overcome the above challenges, enabling real-time and high-performance MPC for dexterous manipulation tasks.

\section{Complementarity-free  Multi-contact Model}

\subsection{Duality of Optimization-based Contact Model}
We proceed by establishing the dual problem of the QP-based contact model (\ref{equ.primal_prob}). First, define the shorthand notations:
\begin{equation}\label{equ.def_mats}
\small
\begin{aligned}
\boldsymbol{\Tilde{J}}:=
\begin{bmatrix}
    \boldsymbol{J}_{1}^n{-}\mu_1\boldsymbol{J}_{1,1}^d
    \\
    \cdots
\\
\boldsymbol{J}_{1}^n{-}\mu_{1}\boldsymbol{J}_{1,n_d}^d\\
\vdots\\
\boldsymbol{J}_{n_c}^n{-}\mu_{n_c}\boldsymbol{J}_{n_c,1}^d
    \\
    \cdots
\\
\boldsymbol{J}_{n_c}^n{-}\mu_{n_c}\boldsymbol{J}_{n_c,n_d}^d
\end{bmatrix},
\,\,\,\,
\boldsymbol{\Tilde{\phi}}:=\begin{bmatrix}
    \phi_{1}
    \\
    \cdots
\\
\phi_{1}\\
\vdots\\
\phi_{n_c}
    \\
    \cdots
\\
\phi_{n_c}
\end{bmatrix},
\,\,\,\,
\boldsymbol{{\beta}}:=\begin{bmatrix}
    \beta_{1,1}
    \\
    \cdots
\\
\beta_{1,n_d}\\
\vdots\\
\beta_{n_c, 1}
    \\
    \cdots
\\
\beta_{n_c, n_d}
\end{bmatrix}
\end{aligned}
\end{equation}
of dimension  $\boldsymbol{\Tilde{J}}\in\mathbb{R}^{n_cn_d\times (n_o\text{+}n_r)}$, $\boldsymbol{\Tilde{\phi}}\in\mathbb{R}^{n_cn_d}$, and $\boldsymbol{{\beta}}\in\mathbb{R}^{n_cn_d}$. Here, $\boldsymbol{\beta}$ is the stacked vector of the dual variables for the $n_cn_d$ constraints in (\ref{equ.primal_prob}). From convex optimization \cite{boyd2004convex}, it can be shown  that the dual problem   of  (\ref{equ.primal_prob}) is
\begin{equation}\label{equ.dual_prob}
    \max_{\boldsymbol{\beta}\geq \boldsymbol{0}}\quad  {-}\frac{1}{2h^2} (h\boldsymbol{b}+\boldsymbol{\Tilde{J}}\tran\boldsymbol{\beta})\tran \boldsymbol{Q}^{-1}(h\boldsymbol{b}+\boldsymbol{\Tilde{J}}\tran\boldsymbol{\beta})-\frac{1}{h}\boldsymbol{\Tilde{\phi}}\tran\boldsymbol{\beta}.
\end{equation}
The optimal primal solution $\boldsymbol{v}^+$ to (\ref{equ.primal_prob}) and the dual solution $\boldsymbol{\beta}^+$ to (\ref{equ.dual_prob}) has the following relationship
\begin{equation}\label{equ.primal_dual_sol}
\boldsymbol{v}^+=\frac{1}{h^2}\boldsymbol{Q}^{-1}(h\boldsymbol{b}+\boldsymbol{\Tilde{J}}\tran\boldsymbol{\beta}^+).
\end{equation}
As pointed out in \cite{todorov2012mujoco}, the dual optimization (\ref{equ.dual_prob}) can be physically interpreted as finding the contact impulse \(\boldsymbol{\beta}\)  that  minimizes the kinetic energy  in  contact frame:  $\frac{1}{2}  {\boldsymbol{v}^+}\tran (h^2\boldsymbol{Q})\boldsymbol{v}^++\frac{1}{h}\boldsymbol{\Tilde{\phi}}\tran\boldsymbol{\beta}$.

It is noted that the dual solution to (\ref{equ.dual_prob}) is not unique because the quadratic matrix $ \boldsymbol{\Tilde{J}} \boldsymbol{Q}^{-1} \boldsymbol{\Tilde{J}}\tran$ is not necessarily of full rank. In MuJoCo  \cite{todorov2012mujoco}, to guarantee the invertibility of the contact model, i.e., the uniqueness of the contact force $\boldsymbol{\beta}$, it adds a small regularization term $\boldsymbol{R}$ to the quadratic matrix, turning (\ref{equ.dual_prob}) into a strongly concave program
\begin{multline}\label{equ.dual_prob_relax}
        \max_{\boldsymbol{\beta}\geq \boldsymbol{0}}\quad  -\frac{1}{2h^2} \boldsymbol{\beta}\tran \left(
    \boldsymbol{\Tilde{J}} \boldsymbol{Q}^{-1} \boldsymbol{\Tilde{J}}\tran+\boldsymbol{R}
    \right) \boldsymbol{\beta}\\-\frac{1}{h}(\boldsymbol{\Tilde{J}} \boldsymbol{Q}^{{-}1}\boldsymbol{b}+\boldsymbol{\Tilde{\phi}})\tran\boldsymbol{\beta}-\frac{1}{2} \boldsymbol{b}\tran\boldsymbol{Q}^{{-}1}\boldsymbol{b}.
\end{multline}
where $\boldsymbol{R}\in\mathbb{R}^{n_cn_d\times n_cn_d}$ is a diagonal regularization matrix with positive entries. The following lemma states the optimality condition for the regularized dual problem (\ref{equ.dual_prob_relax}).
\begin{lemma}\label{lemma.dual}
    The  dual solution to the regularized dual problem (\ref{equ.dual_prob_relax}) satisfies the following dual complementarity constraints:
        \begin{equation}\label{equ.dual_prob_relax_sol}
        \boldsymbol{0}\leq  \boldsymbol{\beta} \perp
        \frac{1}{h} \left(
    \boldsymbol{\Tilde{J}} \boldsymbol{Q}^{-1} \boldsymbol{\Tilde{J}}\tran{+}\boldsymbol{R}
    \right)\boldsymbol{\beta} 
    +\left(\boldsymbol{\Tilde{J}} \boldsymbol{Q}^{{-}1}\boldsymbol{b}+{\boldsymbol{\Tilde{\phi}}}\right)\geq \boldsymbol{0}.
        \end{equation}
\end{lemma}
\noindent
The proof of Lemma \ref{lemma.dual} is in Appendix. This lemma indicates  solving (\ref{equ.dual_prob_relax})  involves  managing complementarity constraints.

\subsection{New Complementarity-Free Multi-Contact Model}
To circumvent the dual complementarity in (\ref{equ.dual_prob_relax_sol}), we propose a new contact model based on Lemma \ref{lemma.dual}. Assume, for now,  the inverse of the first matrix on the right side of (\ref{equ.dual_prob_relax_sol}) can be replaced by a positive definite \emph{diagonal} matrix \(\boldsymbol{K}(\boldsymbol{q})\in\mathbb{R}^{n_cn_d\times n_cn_d}\), i.e.,
        \begin{equation}\label{equ.key_1}
         \boldsymbol{K}(\boldsymbol{q}) \approx(
    \boldsymbol{\Tilde{J}} \boldsymbol{Q}^{-1} \boldsymbol{\Tilde{J}}\tran{+}\boldsymbol{R}
    )^{-1}.
        \end{equation}
We will provide the physical justification for this assumption in the next subsection.
The following lemma states a closed-form solution to the dual complementarity (\ref{equ.dual_prob_relax_sol}) with the replacement in (\ref{equ.key_1}).
\begin{lemma}\label{lemma.key}
Let a positive definite diagonal matrix $\boldsymbol{K}(\boldsymbol{q})$ replace $(
    \boldsymbol{\Tilde{J}} \boldsymbol{Q}^{-1} \boldsymbol{\Tilde{J}}\tran{+}\boldsymbol{R}
    )^{-1}$ as in (\ref{equ.key_1}). The dual complementarity (\ref{equ.dual_prob_relax_sol}) then has a closed-form solution:
\begin{equation}\label{equ.key_dual}
    \boldsymbol{\beta}^+= \max\left(
    -h\boldsymbol{K}(\boldsymbol{q})\big(\boldsymbol{\Tilde{J}} \boldsymbol{Q}^{{-}1}\boldsymbol{b}+{\boldsymbol{\Tilde{\phi}}}\big), \boldsymbol{0}
    \right).
\end{equation}
Furthermore, 
 the approximate  primal solution to (\ref{equ.primal_prob}) is
\begin{equation}\label{equ.key_primal}
      \boldsymbol{v}^+ 
             =
        {\frac{1}{h}\boldsymbol{Q}^{-1}\boldsymbol{b}}+ 
{\frac{1}{h}\boldsymbol{Q}^{-1}\boldsymbol{\Tilde{J}}\tran
        \max\left(
    {-}\boldsymbol{K}(\boldsymbol{q})\big(\boldsymbol{\Tilde{J}} \boldsymbol{Q}^{{-}1}\boldsymbol{b}+{\boldsymbol{\Tilde{\phi}}}\big), \boldsymbol{0}
    \right)}.
\end{equation}
\end{lemma}
\noindent
The proof of Lemma \ref{lemma.key} is given in Appendix. Lemma \ref{lemma.key} presents an important result: by substituting $\big( \boldsymbol{\Tilde{J}} \boldsymbol{Q}^{-1} \boldsymbol{\Tilde{J}}\tran{+}\boldsymbol{R} \big)^{-1}$ with a positive diagonal matrix $\boldsymbol{K}(\boldsymbol{q})$, the complementarity constructs in the original dual condition (\ref{equ.dual_prob_relax_sol}) are eliminated. Critically,  this substitution makes the contact time-stepping prediction (\ref{equ.key_primal}) have closed-form solution and free from complementarity. We thus name (\ref{equ.key_primal}) the \emph{complementarity-free multi-contact model}. Although the above treatment  can cause the time-stepping  prediction $\boldsymbol{v}^+$ in (\ref{equ.key_primal}) to differ from the true primal solution of the original QP-based contact model (\ref{equ.primal_prob}), it offers significant computational benefits, particularly for model-based multi-contact optimization, as will be shown later. In fact, $\boldsymbol{K}(\boldsymbol{q})$ has an intuitive physical interpretation, which will be discussed in the next subsection.

While our complementarity-free time stepping (\ref{equ.key_primal}) approximates (\ref{equ.primal_prob}), we do not expect the resolved velocity $\boldsymbol{v}^+$ to strictly satisfy the original dual cone constraints in (\ref{equ.primal_prob})—although it can potentially achieve so by appropriately choosing the matrix $\boldsymbol{K}(\boldsymbol{q})$. {In fact, relaxing the dual cone constraint in (\ref{equ.primal_prob}) provides our model (\ref{equ.key_primal}) with unique advantages that are not available in (\ref{equ.primal_prob}), particularly in mitigating the non-physical artifacts  with (\ref{equ.primal_prob}) (or general Anitescu  models \cite{anitescu2006optimization,anitescu2010iterative})}, as we elaborate in the next subsection and later experiments in Section \ref{sec.sim_dyn_setting}.

    \begin{figure*}[htbp]
  \centering
  \includegraphics[width=0.95\linewidth]{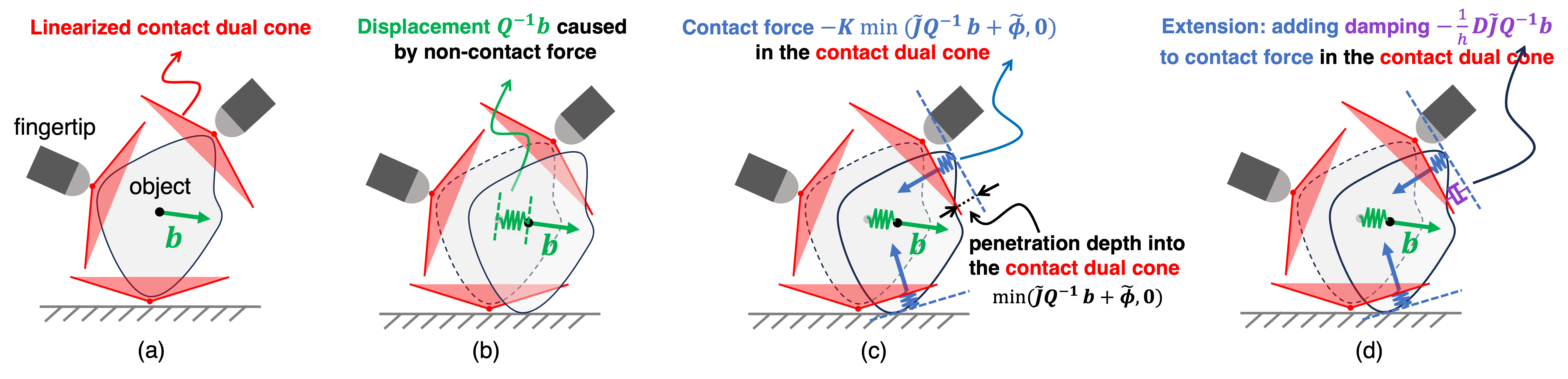}
  \caption{\small The physical interpretation of the proposed complementarity-free multi-contact model (\ref{equ.key_primal}) using a 2D fingertip manipulation example. (a) The dual cone constraints, depicted by red shaded areas, between the object and the ground and fingertips. (b) The displacement (green) caused solely by the non-contact force \(\boldsymbol{b}\) applied to the system to move the object from dashed position to solid position. (c) The spring-like contact force (blue) resulting from the penetration (black) of the contact dual cone. (d) An extension: a damping effect (purple) is added to the contact force term.}
  \label{fig.physics_interpretation}
  \vspace{-10pt}
\end{figure*}
    
   \subsection{Physical Interpretation of the New Model}

We next provide the physical interpretation of the new complementarity-free multi-contact model (\ref{equ.key_primal}) using a 2D fingertip manipulation example  in Fig. \ref{fig.physics_interpretation}. First, recall the linearized   dual cone constraints (\ref{equ.linearized_dual_cone}) in the original QP-based contact model (\ref{equ.primal_prob}). Using (\ref{equ.def_mats}), we can compactly express these  dual cone constraints as
\begin{equation}\label{equ.dual_cone_mat}
h\boldsymbol{\Tilde{J}}\boldsymbol{v}+\boldsymbol{\Tilde{\phi}}\geq \boldsymbol{0}.
\end{equation}
In Fig. \ref{fig.physics_interpretation}(a)-(d), the contact dual cone constraints are shown in red  areas.

Now, returning to the proposed model (\ref{equ.key_primal}), we multiply the time step $h$ on both sides of (\ref{equ.key_primal}), leading to 
\begin{equation}\label{equ.iterpretation1}
h\boldsymbol{v}^+=\boldsymbol{Q}^{-1}\underbrace{\bigg(
\hspace{-5pt}\overset{\substack{\substack{\text{noncontact}\\ \text{force}}\\\downarrow}}{\boldsymbol{b}}
\hspace{-10pt}+ 
\boldsymbol{\Tilde{J}}\tran
        \overbrace{\max\left(
    -\boldsymbol{K}\big(\boldsymbol{\Tilde{J}} \boldsymbol{Q}^{{-}1}\boldsymbol{b}+{\boldsymbol{\Tilde{\phi}}}\big), \boldsymbol{0}
    \right)}^{\text{$:=\boldsymbol{\lambda}_{\text{contact}}$ contact force in contact dual cone}}\bigg)
        }_{\text{total force in generalized coordinate}}.
\end{equation}
Equation (\ref{equ.iterpretation1}) shows that the proposed model (\ref{equ.key_primal}) can be interpreted as a force-spring system, where \(h\boldsymbol{v}^+\) represents the position displacement in the generalized coordinate, and the right side accounts for the total force applied to the system. The matrix \(\boldsymbol{Q}\), defined in (\ref{equ.q_matrix}), encodes the system’s spring stiffness in response to the total force. The total force consists of two components: (i) the non-contact force \(\boldsymbol{b}\) (e.g., gravity and actuation forces), shown in green arrows in Fig. \ref{fig.physics_interpretation}, and (ii) the contact force \(\boldsymbol{\lambda}_{\text{contact}} := \max\big(-\boldsymbol{K}(\boldsymbol{\Tilde{J}} \boldsymbol{Q}^{-1}\boldsymbol{b} + \boldsymbol{\Tilde{\phi}}), \boldsymbol{0}\big)\), shown in blue arrows in Fig. \ref{fig.physics_interpretation}(c), applied within the contact dual cone and then converted to  \(\boldsymbol{\Tilde{J}}\tran\boldsymbol{\lambda}_{\text{contact}}\) in the generalized coordinate. Without contact forces, the system displacement would be \(\boldsymbol{Q}^{-1}\boldsymbol{b}\), as shown in  Fig. \ref{fig.physics_interpretation}(b)-(d)  from the dashed object positions to the solid object positions.

By examining the contact force $\boldsymbol{\lambda}_{\text{contact}}$ in (\ref{equ.iterpretation1}) and combining the contact dual cone constraint in (\ref{equ.dual_cone_mat}), we have
\begin{equation}\label{equ.iterpretation3}
    \boldsymbol{\lambda}_{\text{contact}}
    =-\boldsymbol{K}\hspace{-40pt}\overbrace{\min\Big(\boldsymbol{\Tilde{J}} {\boldsymbol{Q}^{{-}1}\boldsymbol{b}}+{\boldsymbol{\Tilde{\phi}}}, \boldsymbol{0}
    \Big).}^{
    \substack{\text{penetration depth into the contact dual cone (\ref{equ.dual_cone_mat})}\\\text{due to  the displacement $\boldsymbol{Q}^{-1}\boldsymbol{b}$ caused  by non-contact force $\boldsymbol{b}$}}
    }
\end{equation}
Thus,  the contact force \(\boldsymbol{\lambda}_{\text{contact}}\) also behaves like a spring force, proportional to the penetration depth  into the contact \emph{dual cone}, \(\min(\boldsymbol{\Tilde{J}} \boldsymbol{Q}^{-1}\boldsymbol{b} + \boldsymbol{\Tilde{\phi}}, \boldsymbol{0})\), which is  caused by the displacement \(\boldsymbol{Q}^{-1}\boldsymbol{b}\) from the non-contact force \(\boldsymbol{b}\). This has been shown in black arrows in Fig. \ref{fig.physics_interpretation}(c). Notably, the matrix \(\boldsymbol{K}(\boldsymbol{q})\) introduced in Lemma \ref{lemma.key} represents the \emph{stiffness of the contact dual cone}. This  is illustrated in blue in Fig. \ref{fig.physics_interpretation}(c).


As analyzed above, our complementarity-free contact model (\ref{equ.key_primal}) actually leverages the penetration/violation of  the dual cone constraints (\ref{equ.dual_cone_mat}) (also see Fig. \ref{fig.physics_interpretation}(c-d)) to generate spring-like contact force. Thus, (\ref{equ.key_primal}) has relaxed the dual cone constraints. In fact, the dual cone constraints  in (\ref{equ.primal_prob}) are more a mathematical convenience than physical phenomenon (although \cite{anitescu2006optimization,anitescu2010iterative} provide a   microscopic physical interpretation). Strictly enforcing the dual cone constraints will introduce  non-physical artifacts (e.g., boundary layer effect), compared to non-convex complementary models (NCP/LCP) \cite{stewardtrinklePolyhedron,stewart2000implicit,stewart1996implicit}. By relaxation of dual cone constraints, our model (\ref{equ.key_primal}) will reduce such artifacts, as we will show in the later experiments in Section \ref{sec.sim_dyn_setting}.





\subsection{Property and Differentiability of the New Model}

\smallskip
\subsubsection{Automatic satisfaction of Coulomb's Friction Law}    (\ref{equ.iterpretation3}) and (\ref{equ.key_dual}) indicate ${h}\boldsymbol{\Tilde{J}}\tran\boldsymbol{\lambda}_{\text{contact}}=\boldsymbol{\Tilde{J}}\tran\boldsymbol{\beta}^+$. Extending the right side based on the definitions  (\ref{equ.def_mats})  leads to
\begin{equation}\label{equ.contact_force2}
{h}\boldsymbol{\Tilde{J}}\tran\boldsymbol{\lambda}_{\text{contact}}{=}\sum_{i=1}^{n_c}\bigg(
{\boldsymbol{J}_i^n}\tran\sum_{j=1}^{n_d}\beta^+_{i,j}+\mu_i\sum_{j=1}^{n_d} {\boldsymbol{J}_{i,j}^d}\tran\beta^+_{i,j}
\bigg).
\end{equation}
From (\ref{equ.contact_force2}), one can see that at the contact location  $i$, the contact impulse ${h}\boldsymbol{\lambda}_{i, \text{contact}}$ can be decomposed into 
\begin{equation}\label{equ.contact_components}
    \begin{aligned}
        \text{normal force:} \quad h\boldsymbol{\lambda}_{i, \text{contact}}^n&:=\big(\sum\nolimits_{j=1}^{n_d}\beta^+_{i,j}\big)\boldsymbol{n}_i,\\
        \text{frictional force:} \quad h\boldsymbol{\lambda}^d_{i, \text{contact}}&:= \mu_i\sum\nolimits_{j=1}^{n_d} \beta^+_{i,j} \boldsymbol{d}_{i,j}.
    \end{aligned}
\end{equation}
By the triangle inequality,  $\mu_i||\boldsymbol{\lambda}_{i, \text{contact}}^n||\geq ||\boldsymbol{\lambda}_{i, \text{contact}}^d||$ follows. Thus, the contact force $\boldsymbol{\lambda}_{\text{contact}}$ (\ref{equ.iterpretation1}) or (\ref{equ.iterpretation3}) in the proposed model (\ref{equ.key_primal}) automatically satisfies the  Coulomb friction law.



\smallskip
\subsubsection{Differentiability} The proposed complementarity-free model (\ref{equ.key_primal}) is not differentiable due to  $\max$ operation. One thus can replace  $\max$ operation with smooth $\softplus$ function 
\begin{equation}\label{equ.softplus}
    \softplus(x)= \ln(1+e^{\gamma x})/\gamma \quad \text{with}\quad 
    \gamma>0,
\end{equation}
with $\gamma$ controlling its accuracy to $\max(x,0)$. Thus, a differentiable version of the complementarity-free  model (\ref{equ.key_primal}) is 
\begin{equation}\label{equ.key_primal2}
        \boldsymbol{v}^{\text{+}}     =
        {\frac{1}{h}\boldsymbol{Q}^{{-}1}\boldsymbol{b}}+ 
{\frac{1}{h}\boldsymbol{Q}^{-1}\boldsymbol{\Tilde{J}}\tran
        \softplus\left(
    {-}\boldsymbol{K}(\boldsymbol{q})\big(\boldsymbol{\Tilde{J}} \boldsymbol{Q}^{{-}1}\boldsymbol{b}{+}{\boldsymbol{\Tilde{\phi}}}\big)
    \right)}.
\end{equation}

\subsection{Extension to  Dynamic Settings}\label{sec.extension_damping}
Since the contact force $\boldsymbol{\lambda}_{\text{contact}}$ in (\ref{equ.iterpretation1}) is a spring-like force, a natural extension is to include a damping term to stabilize the contact resolution, as shown in Fig. \ref{fig.physics_interpretation}(d), i.e.,
\begin{equation}\label{equ.dampling_contact_force}
\boldsymbol{\lambda}_{\text{contact}}^{\text{ext}} := \max\big(-\boldsymbol{K}(\boldsymbol{q})(\boldsymbol{\Tilde{J}} \boldsymbol{Q}^{-1}\boldsymbol{b} + \boldsymbol{\Tilde{\phi}})\underbrace{-
    \boldsymbol{D}(\boldsymbol{q})\boldsymbol{\Tilde{J}}\big(\boldsymbol{Q}^{{-}1}\boldsymbol{b}/h\big)}_{\text{damping term}}, \boldsymbol{0}\big)
\end{equation}
where $\boldsymbol{D}(\boldsymbol{q})$ is a diagonal damping matrix, and $\frac{1}{h} \boldsymbol{Q}^{{-}1}\boldsymbol{b}$ is the velocity at the contact frame caused  by the non-contact force $\boldsymbol{b}$.
Consequently, the extended complementarity-free multi-contact model  becomes
\begin{equation}\label{equ.key_extension}
        \boldsymbol{v}^+     =
        {\frac{1}{h}\boldsymbol{Q}^{-1}\boldsymbol{b}}+ 
\frac{1}{h}\boldsymbol{Q}^{-1}\boldsymbol{\Tilde{J}}\tran
        \boldsymbol{\lambda}_{\text{contact}}^{\text{ext}}.
\end{equation}
This extension can be less significant in quasi-dynamic systems, because the  velocities are small. However, it plays a significant role in stabilizing contact resolution in dynamic simulation. In Section \ref{sec.sim_dyn_setting}, we will demonstrate the behavior of the contact simulation of (\ref{equ.key_extension}) in dynamic scenarios. A comprehensive investigation of the full dynamic complementarity-free multi-contact model, however, is left for future work.

\section{Complementarity-free Contact-implicit MPC}

With the proposed complementarity-free multi-contact  model, the  MPC formulation (\ref{equ.generic_mpc}) becomes
\begin{equation}\label{equ.cf_mpc}
\begin{aligned}
\min_{\boldsymbol{u}_{0:T{-}1}\in[\mathbf{u}_{\text{lb}}, \mathbf{u}_{\text{ub}}]} \quad 
&
\sum\nolimits_{t=0}^{T-1} \textit{c}\left ( \boldsymbol{q}_t, \boldsymbol{u}_t \right )+\textit{V}\left (\boldsymbol{q}_T\right)
\\ \text{subject to} \quad &
\boldsymbol{q}_{t+1} = \boldsymbol{q}_t\oplus h\boldsymbol{v}_t^+, \quad t=0,...,T{-}1
\\ &
\boldsymbol{v}_t^+ \,\, \text{is  (\ref{equ.key_primal2})  with } 
\boldsymbol{\Tilde{J}} \,\,\text{and} \,\,\boldsymbol{\Tilde{\phi}},\quad
\text{given} \ \boldsymbol{q}_0.
\end{aligned}
\end{equation}
where $\oplus$ is the integration of system position with velocity.

In   (\ref{equ.cf_mpc}), at each prediction step $t$, the contact Jacobian $\boldsymbol{\Tilde J}(\boldsymbol{q}_t)$ and collision distance $\boldsymbol{\Tilde{\phi}}(\boldsymbol{q}_t)$ should ideally be computed  for the predicted state $\boldsymbol{q}_t$ using a collision detection routine \cite{pan2012fcl}. However, including the collision detection operation inside the MPC optimization (\ref{equ.cf_mpc}) is challenging due to its non-differentiability. Fortunately, in a receding horizon framework, one can  do once collision detection for each encountered real-system state $\boldsymbol{q}_0=\boldsymbol{q}^{\text{real}}_k$ to obtain $\boldsymbol{\Tilde{J}}(\boldsymbol{q}_0)$ and $\boldsymbol{\Tilde{\phi}}(\boldsymbol{q}_0)$, which then remain fixed inside the  MPC prediction.  This is equivalent to linearizing the  contact geometry, which  works well for short-horizon MPC  (i.e., $T$ is small). 
We summarize the complementarity-free contact-implicit MPC in Algorithm \ref{alg.cf_mpc}.

\begin{algorithm2e}[htbp]
	\small 
	\SetKwInput{Parameter}{Hyperparameter}
	\SetKwInput{Initialization}{Initialization}
	\Initialization{Hyperparameter $\boldsymbol{K}$ for the model (\ref{equ.key_primal2})} 
 \smallskip
	\For{MPC rollout step $k=0,1,2,...$}{
            Get current system position $\boldsymbol{q}_k^{\text{real}}=\texttt{env}.\texttt{get\_qpos()}$\;
            \smallskip
            Collision detection  to calculate $\boldsymbol{\Tilde{J}}(\boldsymbol{q}_k^{\text{real}})$ and $\boldsymbol{\Tilde{\phi}}(\boldsymbol{q}_k^{\text{real}})$\;
            \smallskip
            Solve MPC  (\ref{equ.cf_mpc}) using any nonlinear optimization  solver, with   $\boldsymbol{q}_0=\boldsymbol{q}_k^{\text{real}}$,  $\boldsymbol{\Tilde{J}}=\boldsymbol{\Tilde{J}}(\boldsymbol{q}_k^{\text{real}})$ and $\boldsymbol{\Tilde{\phi}}=\boldsymbol{\Tilde{\phi}}(\boldsymbol{q}_k^{\text{real}})$\;
            \smallskip
            Apply the first optimal input: \texttt{env}.\texttt{step }($\boldsymbol{u}^*_0(\boldsymbol{q}_k^{\text{real}})$)\;
        }
	\caption{\small Complementarity-free contact-implicit MPC} \label{alg.cf_mpc}
\end{algorithm2e}

In our implementation below, we will show that  a constant diagonal matrix \(\boldsymbol{K}(\boldsymbol{q})\) surprisingly suffices for all the dexterous manipulation tasks (with different objects) we have encountered. A discussion on the setting of \(\boldsymbol{K}(\boldsymbol{q})\) is provided later.    For the other model parameters, such as $\epsilon$ for the object mass regularization in the quasi-dynamic model (\ref{equ.quasi_dyn}), we follow \cite{pang2023global}. $\gamma=100$ in $\softplus$, and   $n_d=4$ in polyhedral frictional cone.  We use the collision detection routine  in MuJoCo \cite{todorov2012mujoco}, which implements \cite{libccd}, and  solve the nonlinear MPC optimization using CasADi \cite{andersson2019casadi} with the IPOPT solver \cite{wachter2006implementation}. The following experiment results are reproducible using the code at \url{https://github.com/asu-iris/Complementarity-Free-Dexterous-Manipulation}, and video demos are available at \url{https://youtu.be/NsL4hbSXvFg}.

\section{Contact Simulation and Comparison}
\subsection{Quasi-Dynamic Setting}
\begin{figure}[h]
  \centering
  \includegraphics[width=1\linewidth]{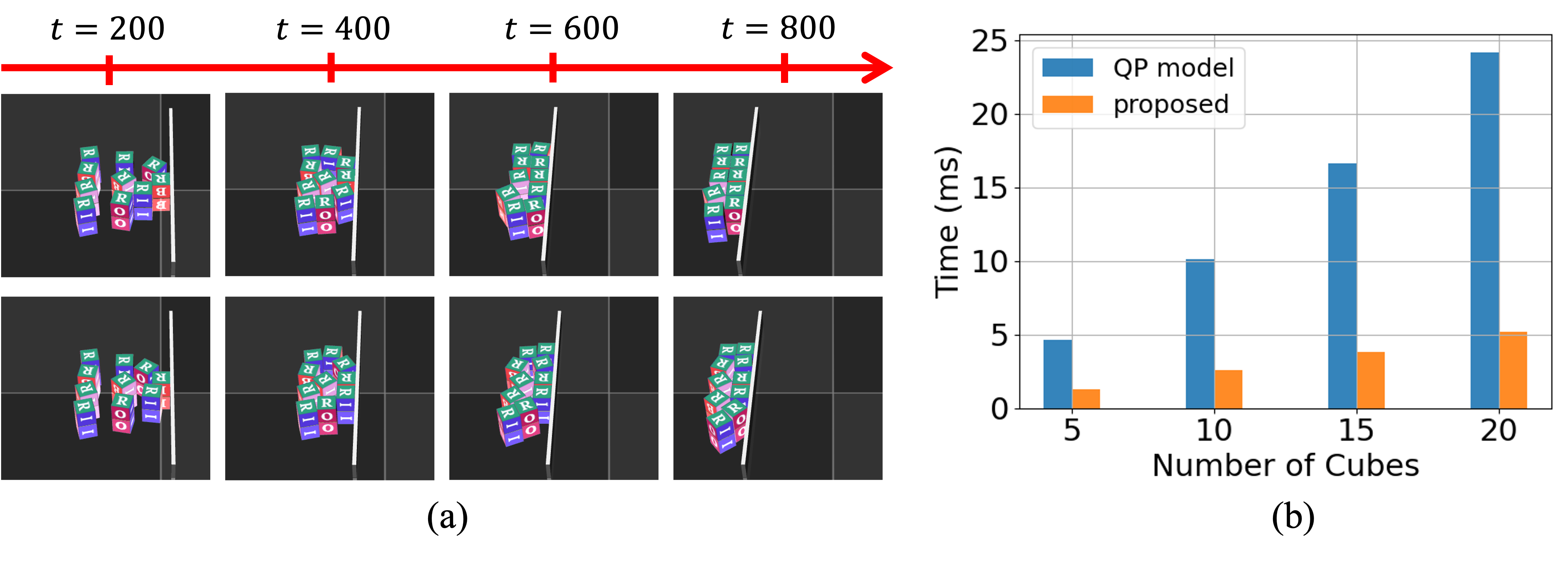}
  \caption{\small (a) The predicted scene of pushing 10 boxes at different steps, by the proposed model (upper row) and QP-based model (bottom row). (b) Timing comparison for one-step prediction. The tests were conducted on a machine with an Apple M2 Pro chip.}
  \label{fig.model_comp}
\end{figure}

We first test the  performance of the  complementarity-free model in predicting a contact-rich pushing-box scene in quasi-dynamic setting, shown in Fig. \ref{fig.model_comp}. In this setup, a bar with a linear joint (moving left-right), actuated by a low-level position controller, pushes varying numbers of cubes on a frictional ground. We compare the proposed model (\ref{equ.key_primal2}) with  QP-based model (\ref{equ.primal_prob}). Since no true quasi-dynamic model exists for benchmarking, the comparison is primarily visual, with the  metric being the timing for time-stepping calculation. The model settings are in Table \ref{table.params}.

{\renewcommand{\arraystretch}{1.2}
\begin{table}[h]
\begin{center}
\begin{threeparttable}
\caption{System and parameter setting}
\begin{tabular}{|c|c|c|}
     \hline
     Parameter & Proposed model (\ref{equ.key_primal2})& QP-based model (\ref{equ.primal_prob})\\
          \hline
          $\boldsymbol{q}\in\mathbb{R}^{7n_{\text{cube}}+1}$ & \multicolumn{2}{c|}{ poses of all cubes and pusher bar (robot) x position} \\
          \hline
          $\boldsymbol{u}\in\mathbb{R}$ & \multicolumn{2}{c|}{desired bar displacement, sent to low-level controller} \\
                \hline
      $h$ & \multicolumn{2}{c|}{0.02 [s]}  \\
     \hline
     $\epsilon$ & \multicolumn{2}{c|}{40}   \\
     \hline
     $\boldsymbol{K}_r$ & \multicolumn{2}{c|}{500}   \\
     \hline
     $\boldsymbol{K}(\boldsymbol{q})$ & $\boldsymbol{I}$ & NA \\
     \hline
\end{tabular}
\label{table.params}
\end{threeparttable}
\end{center}
\vspace{-5pt}
\end{table}
}

Although in the proposed model (\ref{equ.key_primal2}), $\boldsymbol{K}(\boldsymbol{q})$ is configuration-dependant, we here and in the sequel  simply set it as diagonal matrix with identical entries. Our later experiments show that this setting is sufficient for decent contact-rich prediction. Further discussion on setting \(\boldsymbol{K}(\boldsymbol{q})\) is provided later. The QP-based contact model (\ref{equ.primal_prob}) uses  \texttt{OSQP} solver \cite{stellato2020osqp} for time-stepping.
For both models, we initialize the environment with the same random $\boldsymbol{q}_0$ and set the same input  $\boldsymbol{u}_t=-0.001$. The prediction horizon is $T=1000$ steps. The predicted scenes of pushing 10 boxes at different time steps are shown in Fig. \ref{fig.model_comp}(a). To compare the  timing of  one-step prediction, we varied the number of cubes $n_{\text{cube}}$, and the results are given in Fig. 
 \ref{fig.model_comp}(b).

Visually, Fig. \ref{fig.model_comp}(a) shows that the proposed model and the QP-based  model yield similar predictions. However, Fig. \ref{fig.model_comp}(b) shows that the proposed model is 5x faster in one-step prediction. This is expected as the proposed model has an explicit time-stepping formulation, unlike the QP-based contact model, which requires solving a quadratic program at each time step. 

\subsection{Full Dynamic Setting}\label{sec.sim_dyn_setting}

Next, we will evaluate the simulation performance of the extended complementarity-free contact model (\ref{equ.key_extension}) in full dynamic settings, and compare it against several baselines, including classical complementarity-based multi-contact models, a naive QP-based  model (\ref{equ.primal_prob}) (in the dynamic form), and the MuJoCo simulator \cite{todorov2012mujoco}. We will demonstrate that the proposed model: (I) mitigates non-physical artifacts commonly observed in QP models (\ref{equ.primal_prob}) and in general Anitescu’s contact models such as the CCP \cite{anitescu2010iterative} and Convex \cite{anitescu2006optimization} models, due to  the relaxation of dual cone constraints; and (II) achieves sufficient smoothness in contact dynamics due to the spring-damper-based contact resolution  (\ref{equ.key_extension}). Evaluation is conducted in three task settings.

In all dynamic settings below, instead of using the quasi-dynamic model (\ref{equ.quasi_dyn}), we employ the full dynamic model with current system position $\boldsymbol{q}$ and velocity $\boldsymbol{v}$:
\begin{equation}
\boldsymbol{M}(\boldsymbol{q})\boldsymbol{v}^+ - \boldsymbol{M}(\boldsymbol{q})\boldsymbol{v}=
h\boldsymbol{\tau}(\boldsymbol{q},\boldsymbol{v}) + \sum\nolimits_{i=1}^{n_c}{\boldsymbol{J}}\tran\boldsymbol{\lambda}_i,
\end{equation}
where $\boldsymbol{M}(\boldsymbol{q})$ is the configuration-dependent inertia matrix, and $\boldsymbol{\tau}$ represents the sum of all non-contact forces, including inertial, Coriolis, gravitational, and external forces. In the form of (\ref{equ.primal_prob}), it follows that $\boldsymbol{Q}=\boldsymbol{M}(\boldsymbol{q})/h^2$ and $\boldsymbol{b}=\boldsymbol{M}(\boldsymbol{q})\boldsymbol{v}/h + \boldsymbol{\tau}(\boldsymbol{q},\boldsymbol{v})$. As discussed in Section \ref{sec.extension_damping}, the extended complementarity-free contact model (\ref{equ.key_extension}) further incorporates a damping term with a diagonal damping matrix $\boldsymbol{D}(\boldsymbol{q})$.

\begin{figure}[h]
  \centering
  \includegraphics[width=0.90\linewidth]{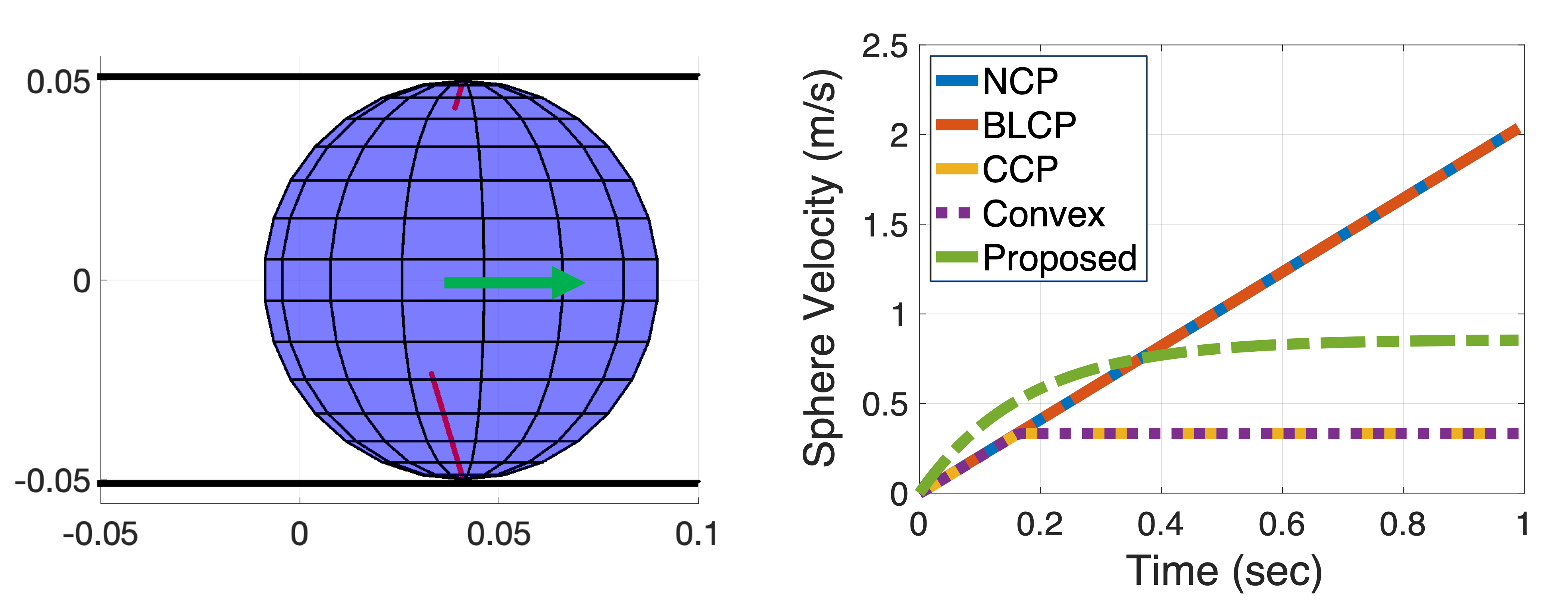}
  \caption{\small  Left: a sphere is accelerated between two frictional planes under an external force (green arrow). Right: the tangential velocity of the sphere versus time. NCP \cite{horak2019similarities,stewardtrinklePolyhedron} and BLCP \cite{erleben2007velocity} contact models  capture the acceleration, while the velocity plateaus for the CCP \cite{anitescu2010iterative} and Convex \cite{anitescu2006optimization} models due to the "sheath effect" induced by the dual cone constraints. The proposed complementarity-free model mitigates such issue because it allows for "violation" of the dual cone constraints while also introducing significant smoothness.}
   \label{fig.full_dyn.setting1}
\end{figure}
\subsubsection{Setting 1: Sphere Sliding Between two Planes} The task follows \cite{horak2019similarities}, which involves sliding a sphere between two
parallel planes as depicted Fig. \ref{fig.full_dyn.setting1}. The sphere with radius of $50$ mm and mass $0.2$kg begins at rest
touching the lower plane. The top plane is $102$mm above the
bottom plane, so initially there is a gap of $2$mm between the
sphere and the top plane. The sphere is only given translational
degrees of freedom and is accelerated with a constant force of
$1$N  in the x-direction (shown in green arrow). The time step is $10$ms, and the friction coefficient is $0.3$. In the extended complementarity-free model (\ref{equ.key_extension}), we set $\boldsymbol{K}(\boldsymbol{q})=20\boldsymbol{I}$ and $\boldsymbol{D}(\boldsymbol{q})=6\boldsymbol{I}$. We compare the proposed (\ref{equ.key_extension}) with NCP \cite{horak2019similarities,stewardtrinklePolyhedron}, BLCP \cite{erleben2007velocity}, CCP \cite{anitescu2010iterative} and Convex \cite{anitescu2006optimization}  contact models. The results are shown in Fig. \ref{fig.full_dyn.setting1}.

With an external force, the sphere is supposed to accelerate with only contact with the lower plane.  NCP \cite{horak2019similarities,stewardtrinklePolyhedron} and BLCP \cite{erleben2007velocity}   capture this.  However,  CCP \cite{anitescu2010iterative} and Convex \cite{anitescu2006optimization} models lead to  the plateaus of velocity, so-called ``sheath effect". This is because  dual cone constraints in CCP and Convex models create the  vertical motion of the sphere, leading to the contact force from the top plane (the red arrows in the left of Fig. \ref{fig.full_dyn.setting1} show the contact impulses).  If the sphere needs to go faster, both planes need to move farther away. In contrast, because our proposed model (\ref{equ.key_extension}) has relaxed the dual cone constraint, our model can effectively mitigate such effect: the plateau velocity of our model is higher than CCP/Convex. Fig. \ref{fig.full_dyn.setting1} also shows the softness of our model. This is expected because our generation of the contact force is handled as a spring-damper system (\ref{equ.key_extension}) in the dual cone (Fig. \ref{fig.physics_interpretation}(d)), which a smooth mapping (\ref{equ.dampling_contact_force}) from system motion.

\begin{figure}[h]
  \centering
  \includegraphics[width=0.99\linewidth]{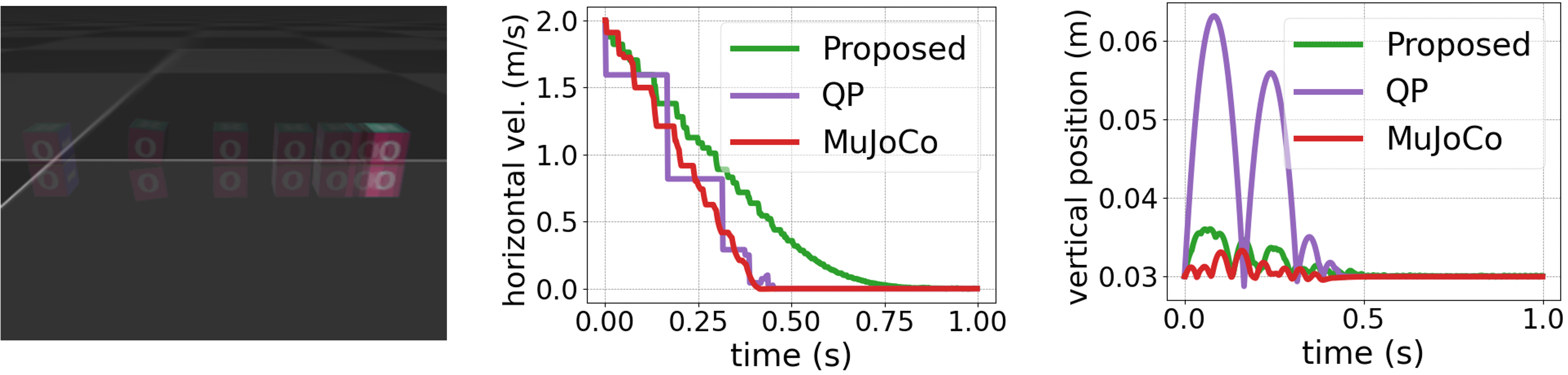}
  \caption{Left: an unactuated box sliding with initial horizontal velocity. Middle and right: the horizontal velocity and vertical position trajectories, respectively. QP-based model (\ref{equ.primal_prob}) exhibits vertical motion artifacts due to the enforcement of dual cone constraints, the proposed model significantly mitigates these effects by relaxing the dual cone constraints. Compared to MuJoCo, our model yields noticeably smoother velocity profiles.}
   \label{fig.full_dyn.setting2}
\end{figure}

\subsubsection{Setting 2: Unactuated Cube Sliding with Initial Horizontal Velocity} In this task, an unactuated cube with a mass of $0.01$ kg, inertia of $6 \times 10^{-6}\boldsymbol{I}$ kg$\cdot\text{m}^2$, and size of $0.06$ m slides on a frictional ground with an initial horizontal velocity of $2.0$ m/s until it comes to rest, as shown in Fig. \ref{fig.full_dyn.setting2}. The friction coefficient is set to $0.5$, and the simulation time step is $2$ ms. In the extended complementarity-free model (\ref{equ.key_extension}), we use $\boldsymbol{K}(\boldsymbol{q}) = \boldsymbol{I}$ and $\boldsymbol{D}(\boldsymbol{q}) = 0.3\boldsymbol{I}$. The baselines include the naive QP-based model (\ref{equ.primal_prob}) (in dynamic form) and the MuJoCo simulator \cite{todorov2012mujoco}. Results are shown in the middle and right panels of Fig. \ref{fig.full_dyn.setting2}. The vertical position trajectory (right) shows that the  QP-based model (\ref{equ.primal_prob}) exhibits vertical motion artifacts due to the enforcement of the  dual cone constraints, whereas the proposed model significantly mitigates these effects by relaxing the dual cone constraints. The horizontal velocity trajectory (middle plot) further demonstrates that, in comparison to MuJoCo, our model yields smoother behavior as a result of the soft handling of contact forces (\ref{equ.dampling_contact_force}).

\begin{figure}[h]
  \centering
  \includegraphics[width=0.99\linewidth]{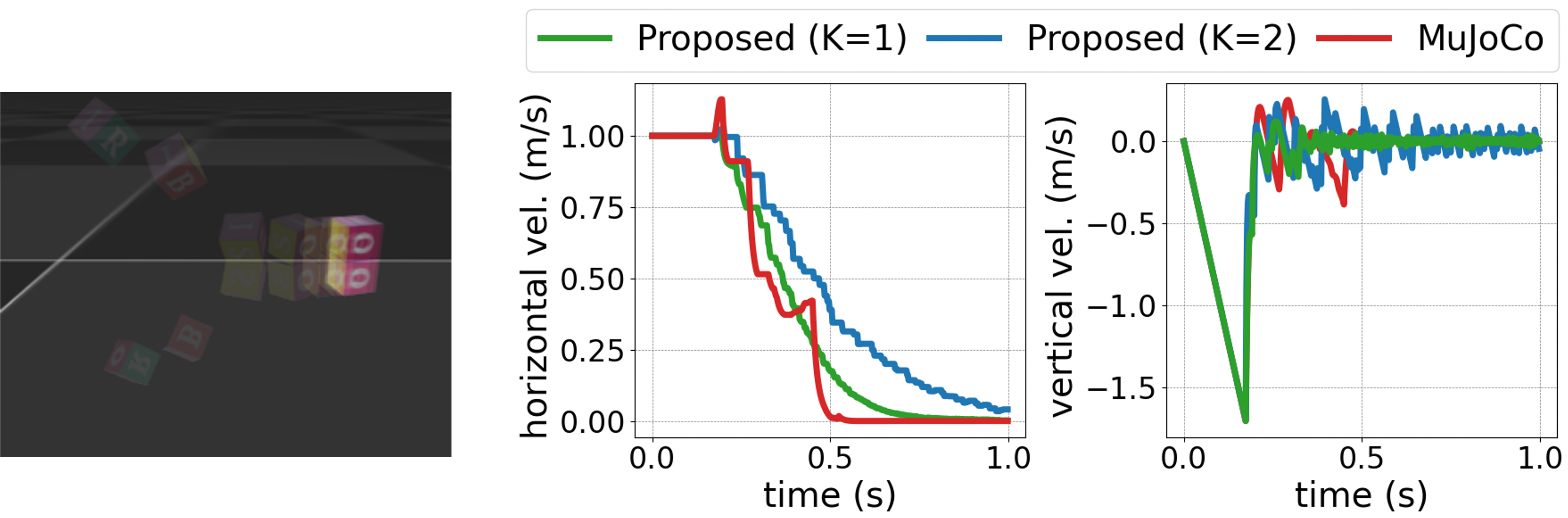}
  \caption{Left: cube free falling, rolling, and sliding on ground. Middle and right: the horizontal  and vertical velocity trajectories, respectively. Compared to MuJoCo, our model yields noticeably smoother velocity profiles. Increasing the stiffness $\boldsymbol{K}$ leads to increased oscillations in the contact resolution.}
   \label{fig.full_dyn.setting3}
\end{figure}
\subsubsection{Setting 3: Cube Free Falling, Rolling, and Sliding on Ground} This task differs from Setting 2 only in the initial pose and velocity of the cube. Specifically, the initial position is $[0.1, 0, 0.2]^\top$, the   quaternion  $[0.577, 0.577, 0.577, 0]^\top$, the  linear velocity  $[1.0, 0, 0]^\top$, and the  angular velocity  $[20, 0, 0]^\top$. These initial conditions cause the cube to undergo free fall, followed by rolling and sliding on the ground until it comes to rest. We evaluate the extended complementarity-free model (\ref{equ.key_extension}) using two different  stiffness matrices: $\boldsymbol{K}(\boldsymbol{q}) = \boldsymbol{I}$ and $\boldsymbol{K}(\boldsymbol{q}) = 2\boldsymbol{I}$, while keeping all other parameters identical to those in Setting 2. The baseline for comparison is the MuJoCo simulator. As shown in Fig. \ref{fig.full_dyn.setting3}, the proposed model produces behavior similar to MuJoCo, but with noticeably smoother velocity profiles. Moreover, increasing the spring stiffness $K$ leads to greater oscillations in the contact response, highlighting the importance of incorporating a damping term in (\ref{equ.key_extension}) for stable and realistic contact behavior.

\section{Fingertips In-Air Manipulation}\label{sec.fingertips_manipulation}

With the quasi-dynamic complementarity-free contact model (\ref{equ.key_primal2}), we evaluate the complementarity-free MPC (Algorithm \ref{alg.cf_mpc}) for three-fingertip manipulation tasks, comparing it to the MPC of the QP-based contact model (\ref{equ.primal_prob}). Hereafter, we refer to the latter as “Implicit MPC” since the QP-based model (\ref{equ.primal_prob}) needs to be converted into its KKT conditions in the MPC  (\ref{equ.cf_mpc}),  similar to \cite{posa2014direct,howell2022trajectory}, to order for the MPC optimization to be solved with existing tools \cite{andersson2019casadi}. All  tasks below use MuJoCo
as  simulation environments. The proposed complementarity-free model (\ref{equ.key_primal2})
is only used in MPC for control optimization.

\subsection{Environment and Task Setup} 
\begin{figure}[h]
  \centering
  \includegraphics[width=0.9\linewidth]{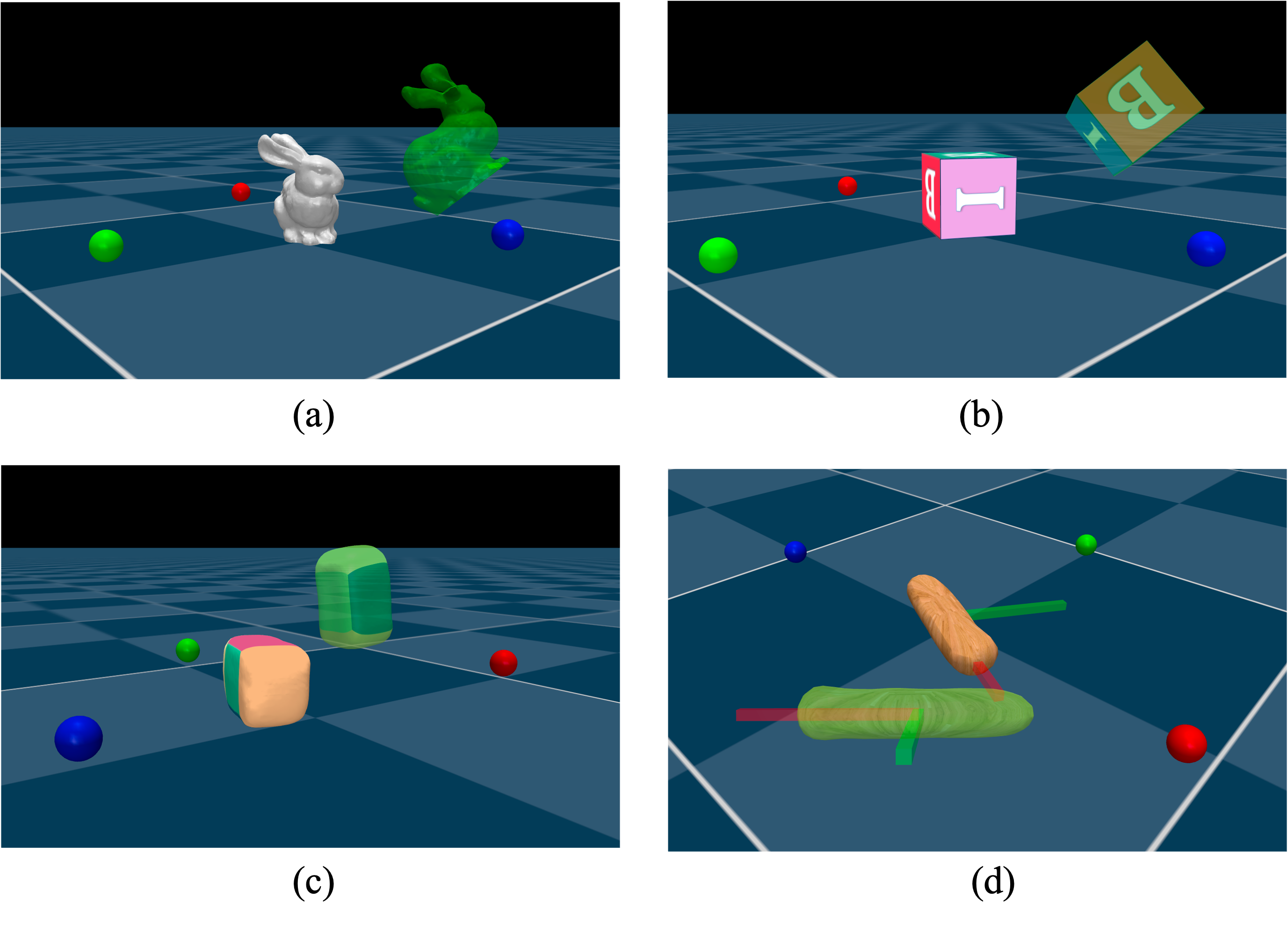}
  \caption{\small Three-fingertip manipulation of different objects. Three fingertips (red, green, blue) are actuated. The tests include 4  objects: (a) Stanford bunny, (b) cube, (c) foam brick, and (d) stick. The  transparent objects are the target poses. For visualization, we have attached a virtual frame to the stick to show its pose. The  diameter of those objects ranges from $0.05$[m] to $0.15$[m].}
  \label{fig.fingertips_setup}
  \vspace{-8pt}
\end{figure}

The three-fingertips manipulation system is shown in Fig. \ref{fig.fingertips_setup}.  MuJoCo \cite{todorov2012mujoco} is used for environment simulation. The three fingertips (red, green, and blue) are actuated with a low-level PD controller ($K_p=100$ and $K_v=2$) with gravity compensation. We use four  different objects: Stanford bunny, cube, foambrick, and stick, all with mass of $0.01$ [kg]. 
The  diameter of those objects ranges from $0.08$ [m] to $0.15$ [m]. The initial pose of objects are statically lying on the ground with random xy position $(x_0^{\text{obj}}, y_0^{\text{obj}})$ and random initial heading (yaw) angle $\psi^{\text{obj}}_0$, uniformly sampled as 
\begin{equation}
    x_0^{\text{obj}}\,\,\, y_0^{\text{obj}}\sim
    \mathbb{U}[-0.025, 0.025]\text{[m]},
    \quad 
    \psi^{\text{obj}}_0 \sim
    \mathbb{U}[-\pi, \pi].
\end{equation}
We will consider three types of manipulation tasks:
\begin{itemize}[leftmargin=*]
    \item \textbf{On-ground rotation:} The target object pose is on the ground. The fingertips only need to move and rotate the object to align it with the target.
    
    \item \textbf{On-ground flipping:} The target requires flipping the object on the ground, with some target poses in non-equilibrium. 
    
    \item \textbf{In-air manipulation:} The target pose is in the air. The fingertips must coordinate to prevent the object from falling while moving it to the target.
\end{itemize}

\subsection{MPC Setting and Results}
\begin{table}[h]
\begin{center}
\begin{threeparttable}
\caption{The model  setting for all objects and  tasks.}
\begin{tabular}{l l}
     \toprule
     Name & Value\\
     \midrule
          $\boldsymbol{q}\in\mathbb{R}^{7+9}$ & object pose and 3D positions of three fingertips 
          \\
          $\boldsymbol{u}\in\mathbb{R}^9$ & {desired fingertip displacement, sent to low-level controller} 
               \\
    $\boldsymbol{K}_r$ & $K_p\boldsymbol{I}$, $K_p{=}100$ is  stiffness of fingertips' low-level control
          \\
     $h$ & $0.1$ [s]
     \\
     $\epsilon \boldsymbol{M}_o/h^2$& $\text{diag}(50,50,50,0.05,0.05, 0.05)$
     \\
             $\boldsymbol{K}(\boldsymbol{q})$ & $\boldsymbol{I}$
     \\
     \bottomrule
\end{tabular}
\label{table.fingertips_model_params}
\end{threeparttable}
\end{center}
\vspace{-8pt}
\end{table}

\begin{figure*}[h]
  \centering
  \includegraphics[width=1\linewidth]{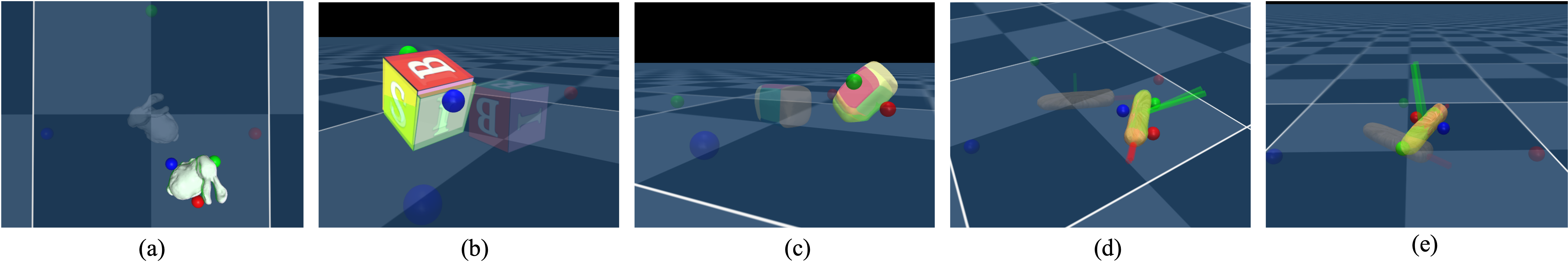}
  \caption{\small On-ground manipulation examples. (a)  On-ground rotation of Stanford bunny. (b)  On-ground flipping of  cube. (c) On-ground flipping of foambrick. (d)-(e) On-ground flipping of stick. In all figures,  the initial object pose and  fingertip positions are shown in transparency, and the final object pose and  fingertip positions are shown in solid. The target object pose is coated with green. \href{https://youtu.be/NsL4hbSXvFg}{Video link}.}
  \label{fig.onground-manipulation-examples}
\end{figure*}

\begin{table*}[t]
\begin{center}
\caption{Results for  on-ground rotation manipulation}
\label{table.fingertips_results_ground_rotation}
\begin{threeparttable}
\begin{tabular}{ccccccccc}
    \toprule
    \multirow{2}{*}{Object} & 
    \multicolumn{2}{c}{{Final position error} (\ref{equ.fingertips_metric})} &
    \multicolumn{2}{c}{{Final heading angle error} (\ref{equ.fingertips_metric})} & 
    \multicolumn{2}{c}{{MPC solving time}} &
    \multicolumn{2}{c}{{Success rate}} 
    \\
    \cmidrule(rl){2-3}
    \cmidrule(rl){4-5} 
    \cmidrule(rl){6-7}
    \cmidrule(rl){8-9}
     &
     Implicit MPC&
     proposed &  
     Implicit MPC&
     proposed &
     Implicit MPC&
     proposed &
     Implicit MPC&
     proposed 
    \\ 
    \midrule
    {\begin{tabular}{@{}l@{}}  Stanford bunny 
    \end{tabular}}
     &        {\begin{tabular}{@{}l@{}}  $0.0146$ [m] \\ $\pm$ $0.0037$  
     \end{tabular}}
    &    \bf{\begin{tabular}{@{}l@{}}  0.0068 [m]\\ $\pm$ 0.0018     \end{tabular}}
    &    {\begin{tabular}{@{}l@{}}  $0.153$ [rad]  \\ $\pm$ $0.065$      \end{tabular}}
    &    \bf{\begin{tabular}{@{}l@{}}  0.0404 [rad] \\ $\pm$ 0.0215      \end{tabular}}
    &    33 [ms]
    &    \textbf{12 [ms]}
    &    95\%
    &     \textbf{100\%}
     \\[10pt]
     {\begin{tabular}{@{}l@{}}  Cube 
     \end{tabular}}
     &              {\begin{tabular}{@{}l@{}} 0.0116 [m] \\ $\pm$ 0.0045      \end{tabular}}          
     &         \bf{\begin{tabular}{@{}l@{}}  0.0102 [m] \\ $\pm$ 0.0047      \end{tabular}}
    &             {\begin{tabular}{@{}l@{}}  0.0939 [rad] \\ $\pm$ 0.0605      \end{tabular}}
    &             \bf{\begin{tabular}{@{}l@{}}  0.0383 [rad] \\ $\pm$ 0.0154      \end{tabular}}
    &    27 [ms]
    &    \textbf{12 [ms]}
    &    100\%
    &\bf{100\%}
     \\[10pt]
     {\begin{tabular}{@{}l@{}}  Foambrick 
     \end{tabular}}
     &             {\begin{tabular}{@{}l@{}}  $0.0137$ [m] \\ $\pm$ $0.0036$      \end{tabular}}     
     &         \bf{\begin{tabular}{@{}l@{}}  0.0079 [m] \\ $\pm$ 0.0027      \end{tabular}}
    &{\begin{tabular}{@{}l@{}}  $0.0938$ [rad] \\ $\pm$ $0.0424$      \end{tabular}}
    &             \bf{\begin{tabular}{@{}l@{}}  0.0454 [rad]\\ $\pm$ 0.0287      \end{tabular}}
    &    47 [ms]
    &    \bf{15 [ms]}
    &    100\%
    &\bf{100\%}
     \\
      \bottomrule
\end{tabular}
\begin{tablenotes}
\item Results for each object are based on 20 random trials, each with random  initial and target object poses. A  trial is  successful if conditions  (\ref{equ.fingertips_termination}) are met \textbf{consecutively} for 20 MPC rollout steps. Final  errors are computed using the last 20 rollout steps in  successful trials. 
\end{tablenotes}
\end{threeparttable}
\end{center}
\vspace{-5pt}
\end{table*}

To show the versatility of the proposed complementarity-free MPC, we  use \emph{the same model parameters and MPC cost function for all tasks and objects}. The model setting is in Table \ref{table.fingertips_model_params}. The MPC path and final cost functions for all  objects and tasks are  defined as
\begin{equation}\label{equ.fingertips_cost_fns}
    \begin{aligned}
        c(\boldsymbol{q}, \boldsymbol{u})&:=
        c_\text{contact}(\boldsymbol{q})+ 0.05c_\text{grasp}(\boldsymbol{q})+ 50\norm{\boldsymbol{u}}^2,
        \\
        V(\boldsymbol{q})&:=
        5000\norm{\mathbf{p}^{\text{obj}}{-}\mathbf{p}_{\text{target}}}^2{+}50\left(1{-}
        (\mathbf{q}_{\text{target}}\tran\mathbf{q}^{\text{obj}})^2
        \right)
    \end{aligned}
\end{equation}
Here, the final cost \(V(\boldsymbol{q})\) defines the “distance-to-goal” for the object’s position and quaternion\footnote[1]{\noindent The quaternion cost/error is defined as $\left(1{-}
        (\mathbf{q}_{\text{target}}\tran\mathbf{q}^{\text{obj}})^2
        \right)$, as used in \cite{huynh2009metrics}. The angle $\theta$ between two normalized quaternions, $\mathbf{q}_1$ and $\mathbf{q}_2$, is calculated as: $\theta=\arccos\left({2(\mathbf{q}_1\tran\boldsymbol{q}_2)^2-1}\right)$. }. In the path cost \(c(\boldsymbol{q}, \boldsymbol{u})\), we define  each term below.
The contact cost term, defined as
\begin{equation}\label{equ.fingertips_contact_cost}
    c_\text{contact}(\boldsymbol{q}):=\sum\nolimits_{i=1}^3\norm{
    \mathbf{p}^{\text{obj}}-\mathbf{p}^{\text{f/t}_i}
    }^2,
\end{equation}
is to encourage the contact between fingertips (position $\mathbf{p}^{\text{f/t}_i}$) and object (position $\mathbf{p}^{\text{obj}}$). The grasp cost term, defined as
\begin{equation}\label{equ.fingertips_grasp_cost}
        c_\text{grasp}(\boldsymbol{q}):=\norm{\mathbf{\Tilde{p}}^{\text{f/t}_1}_{\text{obj}}+\mathbf{\Tilde{p}}^{\text{f/t}_2}_{\text{obj}}+\mathbf{\Tilde{p}}^{\text{f/t}_3}_{\text{obj}}}^2,
\end{equation}
is to encourage  three fingertips to form a stable grasp shape. Here,  $\mathbf{\Tilde{p}}^{\text{f/t}_i}_{\text{obj}}$ is the unit directional vector from the object position $\mathbf{p}^{\text{obj}}$ to  fingertip  position $\mathbf{p}^{\text{f/t}_i}$, viewed in object  frame $R^{\text{obj}}$:
\begin{equation}
\mathbf{\Tilde{p}}^{\text{f/t}_i}_{\text{obj}}:={(R^{\text{obj}})\tran(\mathbf{p}^{\text{f/t}_i}-\mathbf{p}^{\text{obj}})}/{\norm{\mathbf{p}^{\text{f/t}_i}-\mathbf{p}^{\text{obj}}}}.
\end{equation}
This grasp cost is critical for in-air manipulation, where a stable grasp is  essential to prevent  dropping.
The weights for each term in (\ref{equ.fingertips_cost_fns}) are chosen based on the physical unit scale \cite{jin2024task}. The control lower and upper bounds are  $\mathbf{u}_{\text{lb}}=-0.005$ and $\mathbf{u}_{\text{ub}}=0.005$, and  MPC prediction horizon is $T=4$.   In Implicit MPC, we set the complementarity relaxation factor to $5\times 10^{-4}$ for its best performance,  while keeping all other parameters identical to those in the complementarity-free MPC.
 
We deem a manipulation task successful (and terminate the MPC rollout) if both of the following conditions are met:
\begin{equation}\label{equ.fingertips_termination}
\begin{aligned}
    &\norm{\mathbf{p}^{\text{obj}}{-}\mathbf{p}_{\text{target}}}\leq 0.02\,\, \text{[m]},\\
    &1{-}
        (\mathbf{q}_{\text{target}}\tran\mathbf{q}^{\text{obj}})^2\leq 0.015,
\end{aligned}
\end{equation}
\emph{consecutively for 20 MPC rollout  steps.}
A manipulation task is deemed a failure if the object does not satisfy  (\ref{equ.fingertips_termination}) within the maximum MPC rollout length $H = 2000$.

\begin{table*}[t]
\begin{center}
\caption{Results for  on-ground flip manipulation}
\label{table.fingertips_results_ground_flip}
\begin{threeparttable}
\begin{tabular}{ccccccccc}
    \toprule
    \multirow{2}{*}{Object} & 
    \multicolumn{2}{c}{{Final position error} (\ref{equ.fingertips_onground_orierror})} &
    \multicolumn{2}{c}{{Final quaternion error} (\ref{equ.fingertips_onground_orierror})} & 
    \multicolumn{2}{c}{{MPC solving time}} &
    \multicolumn{2}{c}{{Success rate}} 
    \\
    \cmidrule(rl){2-3}
    \cmidrule(rl){4-5} 
    \cmidrule(rl){6-7}
    \cmidrule(rl){8-9}
     &
     Implicit MPC&
     proposed &  
     Implicit MPC&
     proposed &
     Implicit MPC&
     proposed &
     Implicit MPC&
     proposed 
    \\ 
    \midrule
    {\begin{tabular}{@{}l@{}}  Cube 
    \end{tabular}}
    &        {\begin{tabular}{@{}l@{}}  $0.0309$ [m]\\ $\pm$ $0.0254$      \end{tabular}}
    &    \bf{\begin{tabular}{@{}l@{}}  0.0106 [m] \\ $\pm$ 0.0033      \end{tabular}}
    &    {\begin{tabular}{@{}l@{}}  $0.1147$   \\ $\pm$ $0.1549$      \end{tabular}}
    &    \bf{\begin{tabular}{@{}l@{}}  0.0036  \\ $\pm$ 0.0033      \end{tabular}}
    &    31 [ms]
    &    \bf{12 [ms]}
    &    20\%
    &     \bf{100\%}
     \\[10pt]
     {\begin{tabular}{@{}l@{}}  Foambrick 
     \end{tabular}}
     &{\begin{tabular}{@{}l@{}}  $0.0240$ [m] \\ $\pm$ $0.0169$      \end{tabular}}   
     &\bf{\begin{tabular}{@{}l@{}}  0.0101 [m] \\ $\pm$ 0.00322      \end{tabular}}
    &{\begin{tabular}{@{}l@{}}  $0.1389$  \\ $\pm$ $0.2004$      \end{tabular}}
    & \bf{\begin{tabular}{@{}l@{}}  0.0036  \\ $\pm$ 0.0020      \end{tabular}}
    &    57 [ms]
    &    \bf{15 [ms]}
    &    40\%
    &\bf{95\%}
     \\[10pt]
     {\begin{tabular}{@{}l@{}}  Stick 
     \end{tabular}}
     &            {\begin{tabular}{@{}l@{}}  $0.0118$ [m] \\ $\pm$ $0.0156$      \end{tabular}}        
     &         \bf{\begin{tabular}{@{}l@{}}  0.0079 [m] \\ $\pm$ 0.0045      \end{tabular}}
    &{\begin{tabular}{@{}l@{}}  $ 0.0768$  \\ $\pm$ $0.198$      \end{tabular}}
    &             \bf{\begin{tabular}{@{}l@{}}   0.0053 \\ $\pm$ 0.0023      \end{tabular}}
    &    51 [ms]
    &    \bf{15 [ms]}
    &    80\%
    &\bf{100\%}
     \\
      \bottomrule
\end{tabular}
\begin{tablenotes}
\item[]
Each object’s results are based on 20 random trials. 
For  implicit MPC, final position or quaternion errors are computed using all trials due to fewer successful trials. For the proposed method, the errors are computed  using  successful trials.
\end{tablenotes}
\end{threeparttable}
\end{center}
\vspace{-10pt}
\end{table*}


\subsubsection{On-ground Rotation}
In on-ground rotation,  the target object position $\mathbf{p}_\text{target}{=}[x_{\text{target}}, y_{\text{target}}, z_{\text{height}}]\tran$ is  sampled as
\begin{equation}\label{equ.fingertips_onground_target1}
x_{\text{target}} \,\,\,
\text{and} 
\,\,\,
y_{\text{target}} \sim \mathbb{U}[-0.1, 0.1]\,\, \text{[m]},
\end{equation}
and $z_{\text{height}}$ is the height of the object  lying on ground.
The  target object quaternion is $\mathbf{q}_{\text{target}}{=}\texttt{rpyToQuat}(\phi_\text{target}, \theta_\text{target}, \psi_\text{target})$,
with   yaw $\psi_\text{target}$, pitch $\theta_\text{target}$, and roll $\phi_\text{target}$  sampled as
\begin{equation}
    \psi_\text{target}\sim\mathbb{U}[-\pi, \pi], \quad \theta_\text{target}=\phi_\text{target}=0.
\end{equation}

 For each object, we  conduct 20 trials with different random initial and target  poses. The results are given in Table \ref{table.fingertips_results_ground_rotation}, where the manipulation accuracy is evaluated using
\begin{equation}\label{equ.fingertips_metric}
\begin{aligned}
        \text{final position error:}
    \quad 
    &\norm{\mathbf{p}^{\text{obj}}{-}\mathbf{q}_{\text{target}}},
    \\
            \text{final heading angle error:}
    \quad 
    &|\psi_\text{target}-\psi^{\text{obj}}|,
\end{aligned}
\end{equation}
both calculated using the last 20 steps of a MPC rollout. Fig. \ref{fig.onground-manipulation-examples}(a)  shows a manipulation example for the Stanford bunny.

\subsubsection{On-ground Flipping} Here, the random target object position is  sampled from (\ref{equ.fingertips_onground_target1}), and the random target quaternion  $\mathbf{q}_{\text{target}}{=}\texttt{rpyToQuat}(\phi_\text{target}, \theta_\text{target}, \psi_\text{target})$ is sampled by
\begin{equation}\small
    \psi_\text{target}\sim\mathbb{U}[{-}\pi, \pi],  \theta_\text{target}\sim\mathbb{U}[{-}\frac{\pi}{2}, \frac{\pi}{2}], \psi_\text{target}\sim\mathbb{U}[{-}\frac{\pi}{2}, \frac{\pi}{2}].
\end{equation}
Note that some  target poses are in non-equilibrium.

For each object, we  conduct 20 trials with different random initial and target  poses.
The results are  in Table \ref{table.fingertips_results_ground_flip}, where we  quantify the manipulation accuracy by
\begin{equation}\label{equ.fingertips_onground_orierror}
\begin{aligned}
        \text{final position error:}
    \quad 
    &\norm{\mathbf{p}^{\text{obj}}{-}\mathbf{q}_{\text{target}}},
    \\
            \text{final quaternion error\footnote[1]:}
    \quad 
    &1-(\mathbf{q}_{\text{target}}\tran\mathbf{q}^{\text{obj}})^2,
\end{aligned}
\end{equation}
both calculated using the last 20 steps of a MPC rollout.
Fig. \ref{fig.onground-manipulation-examples} visualizes some random trials of on-ground flip manipulation.

\subsubsection{In-air Manipulation} Here, we consider the target object pose in mid-air. Without ground support, the three fingertips must prevent the object from falling while moving it to a target pose. The in-air target object position \(\mathbf{p}_\text{target} = [x_{\text{target}}, y_{\text{target}}, z_{\text{target}}]\tran\) is sampled as
\begin{equation}
\label{equ.fingertips_inair_target_pos}
    \begin{gathered}
    x_{\text{target}}, \,\,
y_{\text{target}} \sim \mathbb{U}[-0.1, 0.1]\,\, [\text{m}],\\
z_{\text{target}}\sim \mathbb{U}[0.03, 0.08] \,\, [\text{m}].
\end{gathered}
\end{equation}
The in-air target object orientation is
\begin{equation}
    \mathbf{q}_{\text{target}}=\texttt{AxisAngleToQuat}(\mathbf{n}_\text{target}, \alpha_\text{target}),
\end{equation}
with random  axis and angle sampled from
\begin{equation}
    \begin{gathered}
        \mathbf{n}_\text{target}\sim\mathcal{N}([0\,\,1\,\,1]\tran, 0.1\boldsymbol{I}),\quad 
        \alpha_\text{target}\sim\mathbb{U}[-\pi, \pi].
    \end{gathered}
\end{equation}

\begin{figure*}[t]
  \centering
  \includegraphics[width=1\linewidth]{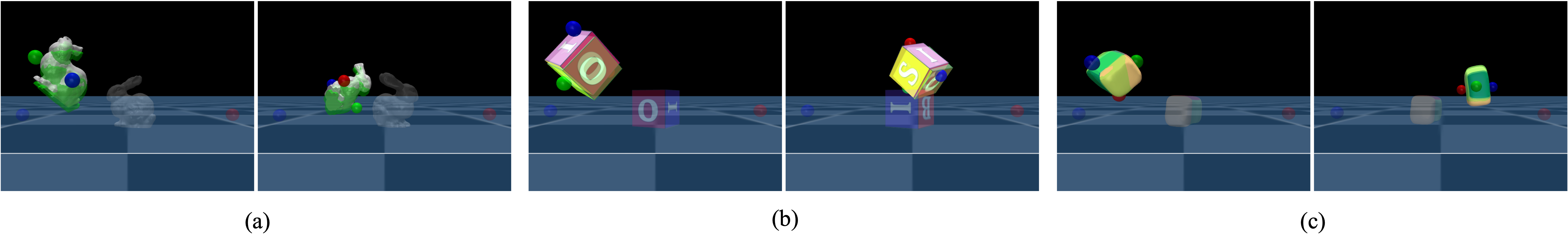}
  \caption{\small In-air manipulation examples. (a) Two random trials of bunny in-air  manipulation. (b) Two random trials of cube in-air   manipulation.  (c) Two random trials of foambrick in-air   manipulation. In all examples,  the initial object pose and  fingertip positions are shown in transparency, and the final object pose and  fingertips positions are in solid. The target object pose is coated in green. \href{https://youtu.be/NsL4hbSXvFg}{Video link}.}
  \label{fig.inair-manipulation}
\end{figure*}

The results of in-air manipulation are listed in Table \ref{table.fingertips_results_inair}. Due to the very low success rate of Implicit MPC for this task type, its results are not included. Manipulation accuracy is measured using the metrics in (\ref{equ.fingertips_onground_orierror}). Fig. \ref{fig.inair-manipulation} shows some random examples of in-air manipulation with different objects.

\begin{table}[h]
\begin{center}
\caption{Results for  in-air manipulation}
\label{table.fingertips_results_inair}
\begin{threeparttable}
\begin{tabular}{ccccc}
    \toprule
    Object 
    & {\begin{tabular}{@{}l@{}}  Final position \\  error (\ref{equ.fingertips_onground_orierror})  \end{tabular}}
      & {\begin{tabular}{@{}l@{}}  Final quaternion\\    error (\ref{equ.fingertips_onground_orierror})  \end{tabular}}
      & {\begin{tabular}{@{}l@{}}  MPC solve-\\ ing time   \end{tabular}} 
      & {\begin{tabular}{@{}l@{}}  Success \\ rate \end{tabular}} 
    \\[5pt] 
    \midrule
    {\begin{tabular}{@{}l@{}}  Stanford  \\ bunny  \end{tabular}}
    & {\begin{tabular}{@{}l@{}}  $0.0082$ [m] \\ $\pm$  $0.0021$  \end{tabular}} 
    & {\begin{tabular}{@{}l@{}}  $ 0.0025$  \\ $\pm$  $0.0026$  \end{tabular}} 
    & 13 [ms]
    &90\%
    \\[10pt]
        Cube
    & {\begin{tabular}{@{}l@{}}  $0.0093$ [m] \\ $\pm$  $0.0021$  \end{tabular}} 
    & {\begin{tabular}{@{}l@{}}  $ 0.0029$  \\ $\pm$  $0.0024$  \end{tabular}} 
    & 13 [ms]
    &90\%
        \\[10pt]
        Foambrick
    & {\begin{tabular}{@{}l@{}}  $0.0074$ [m] \\ $\pm$  $0.0016$  \end{tabular}} 
    & {\begin{tabular}{@{}l@{}}  $ 0.0032$  \\ $\pm$  $0.0021$  \end{tabular}} 
    & 17 [ms]
    &95\%
     \\
      \bottomrule
\end{tabular}
\begin{tablenotes}
\item[] Each object’s results are based on 20 trials, each with a random  initial and target pose. A  trial is considered successful if  conditions  (\ref{equ.fingertips_termination}) are met {consecutively} for 20 MPC rollout steps.
\end{tablenotes}
\end{threeparttable}
\end{center}
\vspace{-20pt}
\end{table}

\subsection{Result Analysis}
On  results in Tables \ref{table.fingertips_results_ground_rotation}-\ref{table.fingertips_results_inair}, we make the following comments.

(1) The proposed complementarity-free MPC consistently outperforms Implicit MPC (i.e., MPC with complementarity model) across various manipulation tasks in terms of   success rate, accuracy, and speed. The superiority is more evident in complex tasks like on-ground flipping and in-air manipulation. 

(2) Quantitatively, the proposed complementarity-free MPC consistently achieves state-of-the-art results, with 100\% success in on-ground rotation, over 95\% in on-ground flipping, and over 90\% in in-air manipulation. The average of the final  object position error is $8.9$ [mm] (noting object diameters range from $50$ [mm]  to $150$ [mm]) and final quaternion error is  0.0035 (equivalent to $6.844^\circ$) across all tasks and objects. The MPC frequency consistently exceeds 58 Hz for all tasks.

(3) The above state-of-the-art results are largely due to the explicit, complementarity-free nature of the proposed contact model, which significantly improves the feasibility of contact optimization. In IPOPT solver, the complementarity-free MPC optimization usually converges to a solution within around 20 iterations. In contrast, Implicit MPC optimization    struggles and takes longer to converge due to the numerous complementarity constraints, even with relaxation.




\subsection{How to Set the Model Parameter $\boldsymbol{K}$?}
\begin{figure}[h]
  \centering
  \includegraphics[width=0.8\linewidth]{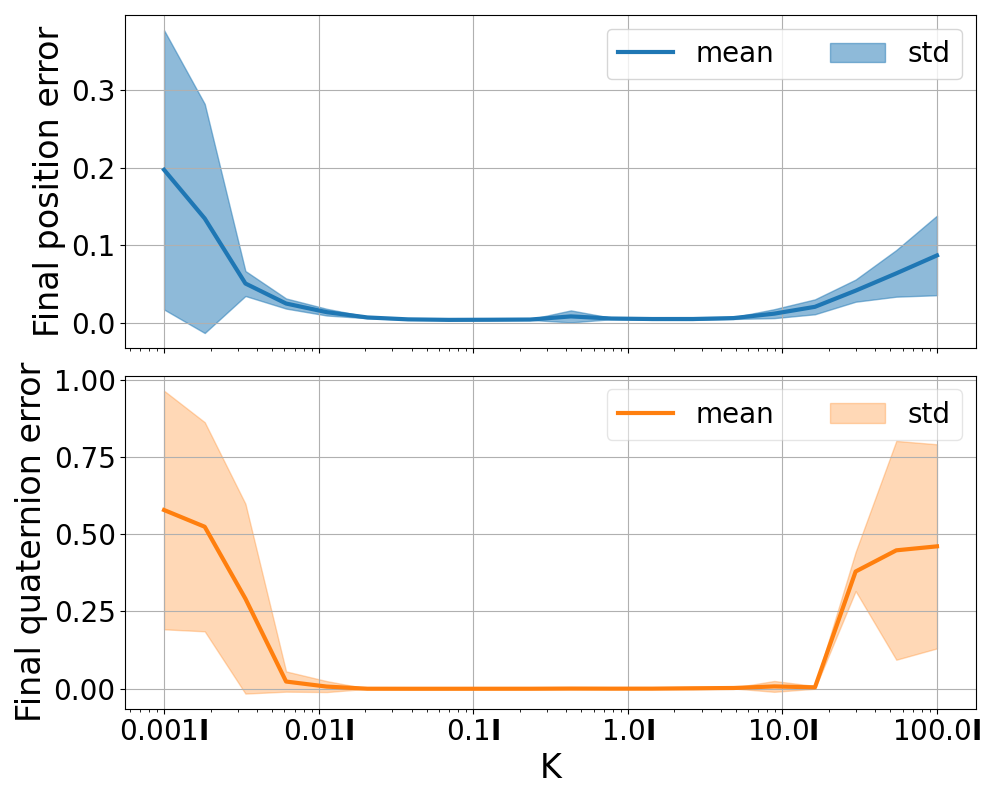}
  \caption{\small Complementarity-free MPC performance with different parameter \(\boldsymbol{K}\) values for on-ground rotation tasks. Shaded areas are the standard deviation, computed from 10 random trials.
  }
  \label{fig.param_setting}
\end{figure}
In the above implementation,  model parameters (Table \ref{table.fingertips_model_params}) follow  standard setting \cite{pang2022TRO}, except for the new stiffness matrix \(\boldsymbol{K}\). We simply set \(\boldsymbol{K}\) as an identity matrix for all tasks. Now, we examine how different \(\boldsymbol{K}\) values affect MPC performance using the bunny on-ground   rotation tasks. All parameters follow Table \ref{table.fingertips_model_params}, except \(\boldsymbol{K}\), which varies from $10^{-3}\boldsymbol{I}$ to $10^{2}\boldsymbol{I}$. For each \(\boldsymbol{K}\), we run 10 trials with different initial and target poses.  MPC rollout length is 500. Fig. \ref{fig.param_setting} shows the task performance  versus different \(\boldsymbol{K}\), where the final position and quaternion errors (\ref{equ.fingertips_onground_orierror}) are calculated at the last rollout step.

Fig. \ref{fig.param_setting} shows that good  performance is achieved across a wide range of \(\boldsymbol{K}\) settings, from \(0.01\boldsymbol{I}\) to \(10\boldsymbol{I}\). Outside this range, MPC performance declines. This highlights the flexibility in choosing an effective \(\boldsymbol{K}\) value. Additionally, since the proposed complementarity-free model is fully differentiable, more complex \(\boldsymbol{K}(\boldsymbol{q})\) settings can be learned from environment data, which we plan to explore in future work.

{\renewcommand{\arraystretch}{1.2}
\begin{table*}[t]
\begin{center}
\begin{threeparttable}
\caption{The model  settings for two dexterous manipulation environments}
\begin{tabular}{l l l}
     \toprule
     Name
        & \textbf{TriFinger in-hand manipulation} & \textbf{Allegro hand on-palm reorientation}\\
     \midrule
          $\boldsymbol{q}$ &   $\boldsymbol{q}\in\mathbb{R}^{7+9}$ including object's 6D pose  and  TriFinger's 9 joints      
          & $\boldsymbol{q}\in\mathbb{R}^{7+16}$ including object's 6D pose  and  all fingers' 16 joints
          \\\hline
          $\boldsymbol{u}$ & $\boldsymbol{u}\in\mathbb{R}^9$: {desired finger joint displacement, sent to low-level  control}
          & $\boldsymbol{u}\in\mathbb{R}^{16}$: {desired finger joint displacement, sent to low-level  control}
          \\\hline
    $\boldsymbol{K}_r$ &  $K_p\boldsymbol{I}$,\,\, $K_p=10$ is the stiffness  of  fingers' low-level controller & $K_p\boldsymbol{I}$,\,\,  $K_p=1$ is the stiffness  of fingers' low-level controller
     \\\hline
     $h$ &  \multicolumn{2}{c}{$h=0.1$ [s]}
     \\\hline
     $\epsilon \boldsymbol{M}_o/h^2$&  \multicolumn{2}{c}{$\text{diag}(50,50,50,0.1,0.1, 0.1)$ for all objects} 
     \\\hline
             $\boldsymbol{K}(\boldsymbol{q})$ & \multicolumn{2}{c}{$0.5\boldsymbol{I}$ for all objects (chosen based on the results in Fig. \ref{fig.param_setting})}
     \\
     \bottomrule
\end{tabular}
\label{table.dexhand_model_params}
\end{threeparttable}
\end{center}
\vspace{-10pt}
\end{table*}
}

\section{Real-time Dexterous In-hand Manipulation}

\begin{figure}[h]
  \centering
  \includegraphics[width=0.9\linewidth]{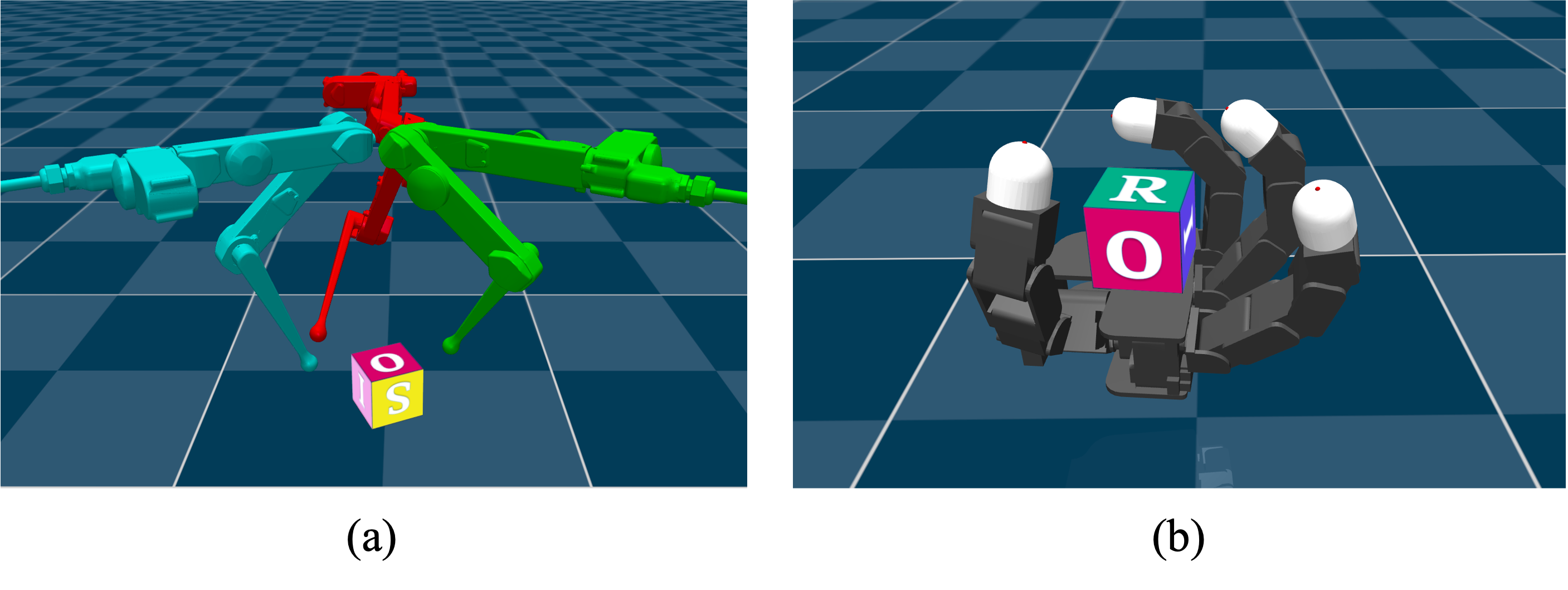}
  \caption{\small  (a) TriFinger in-hand manipulation.  (b) 4-Fingered Allegro hand on-palm reorientation.  Both simulation environments are built using MuJoCo \cite{todorov2012mujoco}.}
  \label{fig.dexman_setup}
\end{figure}

With the quasi-dynamic complementarity-free contact model (\ref{equ.key_primal2}),  the  complementarity-free MPC is evaluated in   dexterous in-hand  manipulation tasks with two environments:
\begin{itemize}[leftmargin=*]
    \item \textbf{TriFinger in-hand manipulation.} As in Fig. \ref{fig.dexman_setup}(a), the three-fingered robotic hand faces down and each finger has 3 DoFs, actuated with low-level joint PD controllers with proportional gain ${K}_p=10$ and damping gain $K_d=0.05$.
    \item \textbf{Allegro hand on-palm reorientation.} An Fig. \ref{fig.dexman_setup}(b), the four-fingered hand faces up and each finger has 4 DoFs. We  implemented a low-level joint PD  controller for each finger with control gains: $K_p=1$ and  $K_d=0.05$.
\end{itemize}
All tasks use MuJoCo
as  simulation environments. The proposed complementarity-free model (\ref{equ.key_primal2})
is only used in MPC for control optimization.

In each environment, we focus on dexterous manipulation of 17 objects with diverse geometries,  shown in Fig. \ref{fig.intro}. The object meshes are from the ContactDB dataset \cite{brahmbhatt2019contactdb}, with some proportional scaling to fit the workspace of the two hands. Object diameters range from 50 mm to 150 mm. In both environments, the goal is to manipulate objects from random initial poses to randomly given target poses. To show the versatility of our complementarity-free MPC, we use the same object-related parameters in our model across  environments and  objects, despite varying physical properties. The model  settings are listed in Table \ref{table.dexhand_model_params}.


\subsection{TriFinger In-hand Manipulation}

\begin{figure*}[t]
  \centering
  \includegraphics[width=0.92\linewidth]{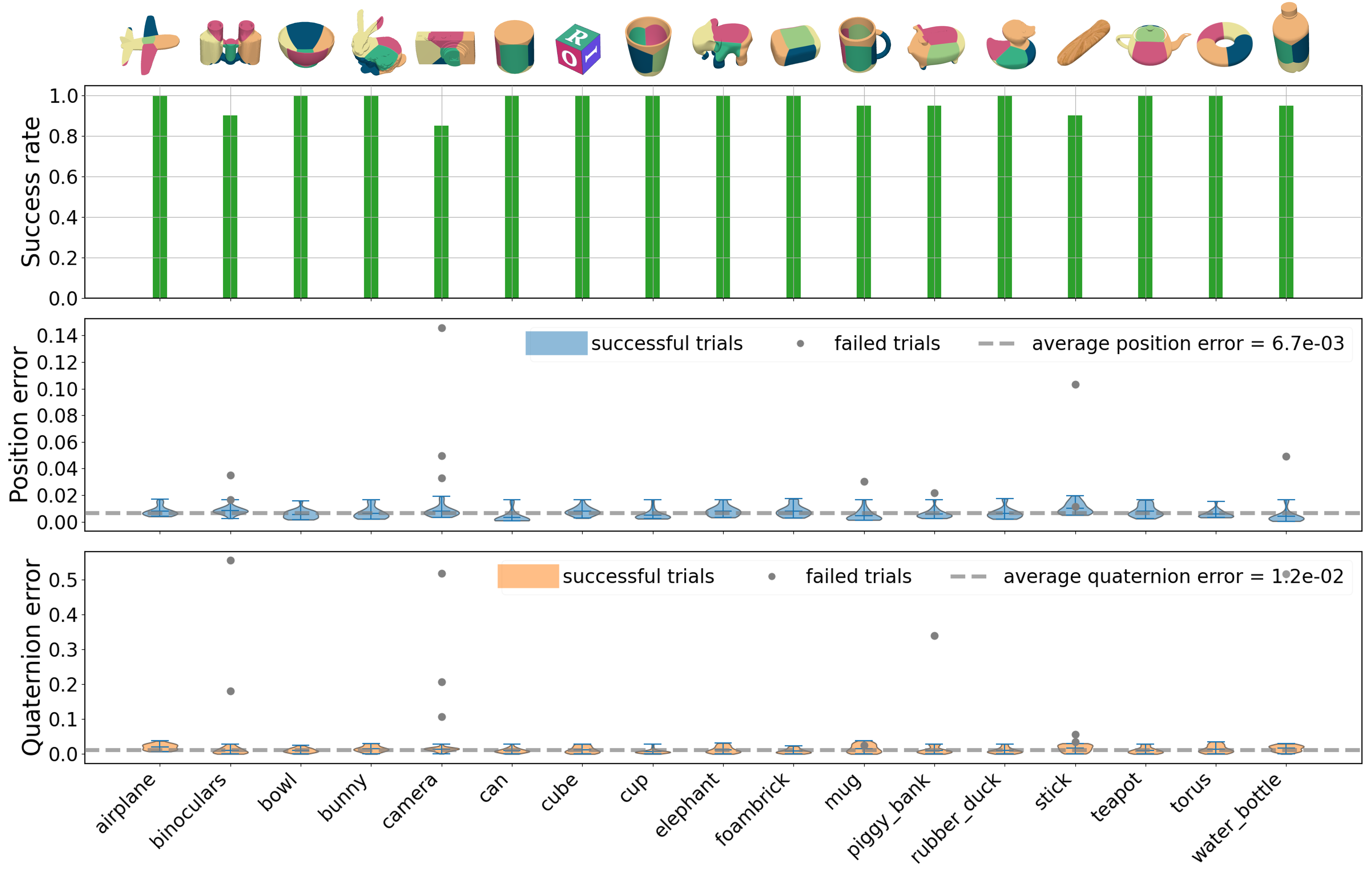}
  \caption{\small Results of the TriFinger in-hand manipulation for various objects. For each object on the x-axis, the upper panel shows the success rate across 20 trials based on criterion (\ref{equ.trifinger_criterion}). The middle and bottom panels show violin plots of the final position and quaternion errors in successful trials, with errors in failed trials  indicated by gray dots. All errors are calculated according to (\ref{equ.fingertips_onground_orierror}). \href{https://youtu.be/NsL4hbSXvFg}{Video link}.
  }
  \label{fig.trifinger_results}
  \vspace{-10pt}
\end{figure*}

\subsubsection{Task Setup} The TriFinger in-hand manipulation task involves moving various objects (the first row of Fig. \ref{fig.intro}) from a random on-ground initial pose \((\mathbf{p}^{\text{obj}}_{0}, \mathbf{q}^{\text{obj}}_{0})\) to a given  random on-ground target pose \((\mathbf{p}_{\text{target}}, \mathbf{q}_{\text{target}})\). The initial object position \(\mathbf{p}^{\text{obj}}_{0} {=} [x_0^\text{obj}, y_0^\text{obj}, z_{\text{height}}]\) is sampled as
\begin{equation}\label{equ.trifinger-init-pos}
    x_0^\text{obj}, \,\, y_0^\text{obj}\sim \mathbb{U}[-0.05, 0.05] \,\, 
    \text{[m]},
\end{equation}
with $z_{\text{height}}$ being the height of an object when it rests on ground. The initial object quaternion is $\mathbf{q}^{\text{obj}}_{0}{=}\texttt{rpyToQuat}(\phi_0^\text{obj},0, 0)$, with its heading (yaw) $\phi_0^\text{obj}$  sampled from 
\begin{equation}\label{equ.trifinger-init-yaw}
    \phi_0^\text{obj}\sim \mathbb{U}[-\frac{\pi}{2}, \frac{\pi}{2}] \,\, 
    \text{[rad]}.
\end{equation}
The target  position $\mathbf{p}_{\text{target}}{=}[x_\text{target},y_\text{target},z_{\text{height}}]$ is sampled as
\begin{equation}\label{equ.trifinger-target-pos}
       x_\text{target}, \,\, y_\text{target}\sim \mathbb{U}[-0.05, 0.05] \,\, 
    \text{[m]},
\end{equation}
 and the target object heading (yaw) is sampled from 
\begin{equation}\label{equ.trifinger-target-yaw}
    \phi_\text{target}\sim \mathbb{U}[-\frac{\pi}{2}, \frac{\pi}{2}] \,\, 
    \text{[rad]}.
\end{equation}
The above target  range is set based on the reachable  workspace of the TriFinger hand, consistent with \cite{jin2024task}.

In the complementarity-free  MPC, the path and final cost functions are defined  as
\begin{equation}\label{equ.trifinger_cost_fns}
    \begin{aligned}
        c(\boldsymbol{q}, \boldsymbol{u})&{:=}
        c_\text{contact}(\boldsymbol{q})+ 0.05c_\text{grasp}(\boldsymbol{q})+ 10\norm{\boldsymbol{u}}^2,
        \\
        V(\boldsymbol{q})&{:=}
        5000\norm{\mathbf{p}^{\text{obj}}{-}\mathbf{p}_{\text{target}}}^2{+}50\left(1{-}
        (\mathbf{q}_{\text{target}}\tran\mathbf{q}^{\text{obj}})^2
        \right),
    \end{aligned}
\end{equation}
respectively. Here, the contact cost term $c_{\text{contact}}$ is defined as the sum of squared distance between object  and  fingertip positions:
\begin{equation}
 c_\text{contact}(\boldsymbol{q}):=\sum\nolimits_{i=1}^3\norm{
    \mathbf{p}^{\text{obj}}-\mathbb{FK}(\boldsymbol{q}^{\text{finger}_i})
    }^2,
\end{equation}
with $\mathbb{FK}(\boldsymbol{q}^{\text{finger}_i})$ the forward kinematics of finger $i$; other cost terms follow  (\ref{equ.fingertips_cost_fns}).
The MPC control bounds are $\mathbf{u}_{\text{lb}}=-0.01$ and $\mathbf{u}_{\text{ub}}=0.01$. MPC prediction horizon is $T=4$.

\smallskip


\begin{figure*}[t]
  \centering
  \includegraphics[width=1\linewidth]{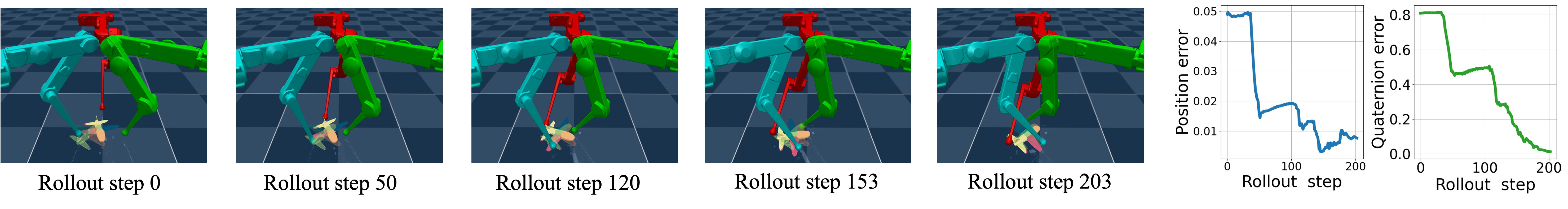}
  \caption{\small An example of the TriFinger in-hand manipulation of an airplane object. The first five panels are screenshots of a MPC rollout at different steps, with the target  shown in transparency. The final panel draws the position and quaternion errors (\ref{equ.fingertips_onground_orierror}) along the rollout steps.
  }
  \label{fig.trifinger_example_success}
\end{figure*}

\subsubsection{Results} For each object, we run the MPC policy for 20 trials. In each trial, we randomize the initial object  and target poses according to (\ref{equ.trifinger-init-pos})-(\ref{equ.trifinger-target-yaw}).  The MPC rollout length is set to $H=500$. The rollout early terminates if  both of the following conditions 
\begin{equation}\label{equ.trifinger_termination}
\begin{aligned}
    &\norm{\mathbf{p}^{\text{obj}}{-}\mathbf{p}_{\text{target}}}\leq 0.02 
    \,\text{[m]}, \,\, 1{-}
        (\mathbf{q}_{\text{target}}\tran\mathbf{q}^{\text{obj}})^2\leq 0.04,
\end{aligned}
\end{equation}
are met consecutively for  20 rollout  steps. We define 
\begin{equation}\label{equ.trifinger_criterion}
    \text{a trial}  
    \begin{cases}
        \text{succeeds,} & \text{if (\ref{equ.trifinger_termination}) is met consecutively for  20}\\[-2pt]
        & \text{rollout steps within $H=500$},\\[2pt]
        \text{fails,} & \text{otherwise}.
    \end{cases}
\end{equation}

Fig. \ref{fig.trifinger_results} presents the TriFinger in-hand manipulation results, with the x-axis representing each object. The upper panel shows the success rate across 20 trials, while the middle and bottom panels respectively show  violin plots of the final position and quaternion errors in successful trials, calculated according to (\ref{equ.fingertips_onground_orierror}). Errors in the failed trials are also shown in ``gray dots". The key performance metrics across all objects and all trials are summarized in Table \ref{table.trifinger_results_summary}.

\begin{table}[h]
\begin{center}
\begin{threeparttable}
\caption{Summary of the TriFinger in-hand manipulation results across all objects and trials}
\begin{tabular}{l l}
     \toprule
     Metric & Value\\
     \midrule
        Overall success rate   & $97.0\% \pm 4.5\%$
          \\
        Overall final position error \tnote{a}  (\ref{equ.fingertips_onground_orierror})   & $0.0067  \pm 0.0041$ \text{[m]}
          \\
                  Overall final quaternion error (\ref{equ.fingertips_onground_orierror})    & $0.0124 \pm 0.0090$ 
          \\
                  Overall final angle error\tnote{b}   & $ 11.78^\circ  \pm 5.10^\circ$ 
          \\
        Average MPC solving time per rollout step  & $13 \pm 3$ \text{[ms]}
          \\
     \bottomrule
\end{tabular}
\label{table.trifinger_results_summary}
\begin{tablenotes}
\item[a]
Note that  object diameters range from 50 [mm] to 150 [mm].
\item[b] The angle error is calculated by $\theta{=}\arccos({2(\mathbf{q}_{\text{target}}\tran\boldsymbol{q}^{\text{obj}})^2{-}1})$. 
\end{tablenotes}
\end{threeparttable}
\end{center}
\vspace{-10pt}
\end{table}

\subsubsection{Analysis}
Fig. \ref{fig.trifinger_results} and Table \ref{table.trifinger_results_summary} demonstrate that the proposed complementarity-free MPC consistently achieves high success rates and accuracy in TriFinger in-hand manipulation across diverse objects, with MPC running at around 80 Hz. These results represent state-of-the-art performance compared to the previous work \cite{jin2024task}. Fig. \ref{fig.trifinger_example_success} illustrates a  MPC rollout example for in-hand manipulation of an airplane object.

\begin{figure}[h]
  \centering
  \includegraphics[width=0.95\linewidth]{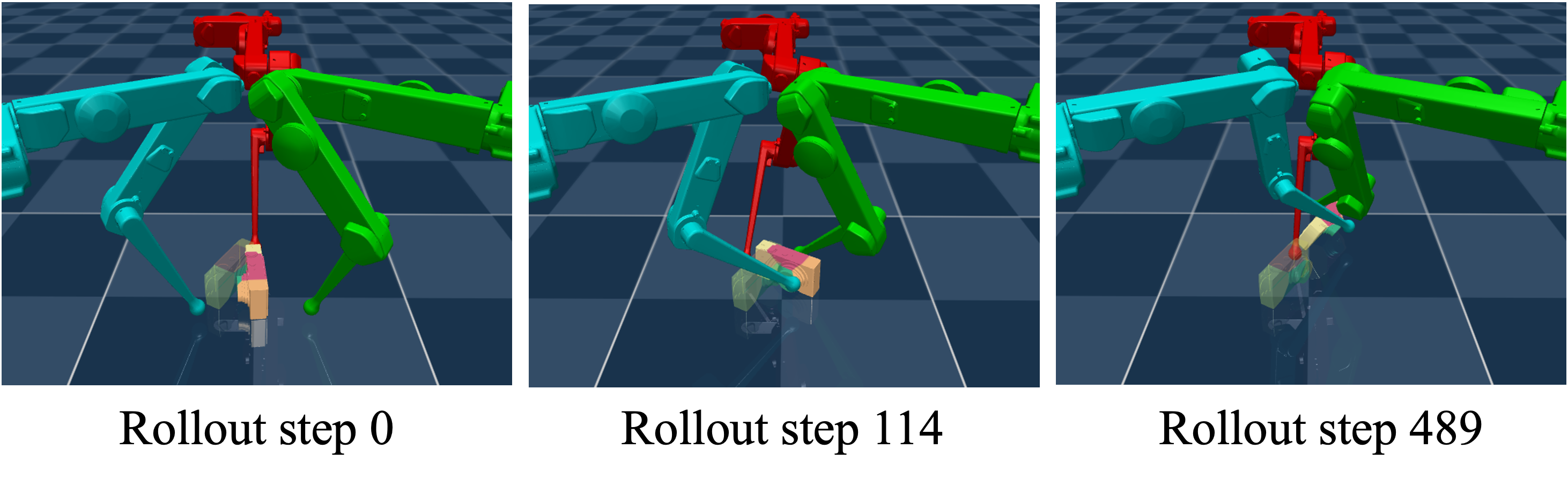}
  \caption{\small An example of failed  trial in manipulating camera.}
  \label{fig.trifinger_example_failure}
\end{figure}

Despite the high overall success rate,  failures occur for few  objects, e.g., camera (success rate $85\%$) and stick (success rate $90\%$). In Fig. \ref{fig.trifinger_example_failure}, we  show  a failed trial. We observed that the primary cause of the manipulation failure was   the self-collision of the three fingers. Since self-collision is not included as a motion constraint in our MPC optimization, it can occur when  object pose approaches the edge of valid workspace.

\subsection{Allegro Hand On-palm Reorientation}

\begin{figure*}[t]
  \centering
  \includegraphics[width=1\linewidth]{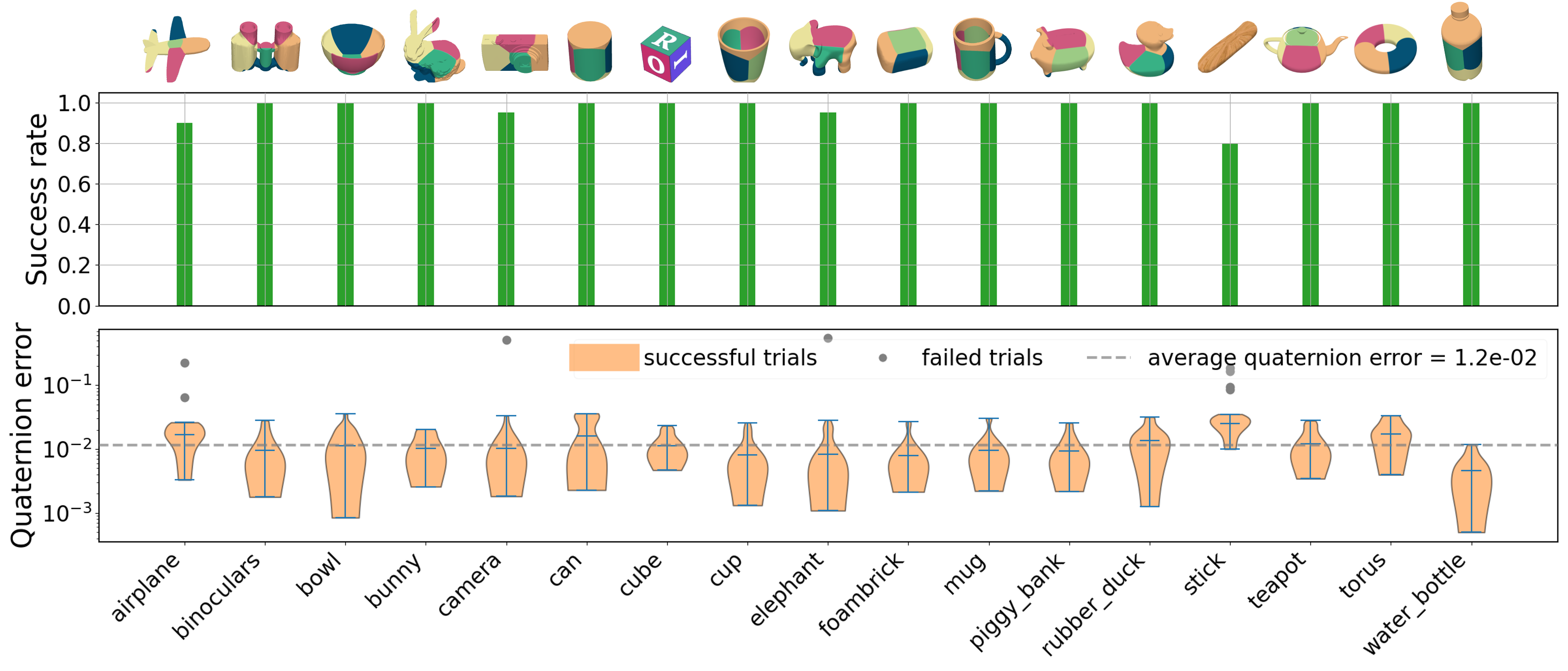}
  \caption{\small Results of the Allegro hand on-palm reorientation of diverse objects. For each object (in x-axis), the upper panel shows the success rate across 20 random trials based on the criterion (\ref{equ.allegro_criterion}). The bottom panel shows the violin plots of  quaternion errors (log-scale) in all successful trials.
  Here,  errors in failed trials are shown as ``gray dots". The  quaternion error is defined  in (\ref{equ.fingertips_onground_orierror}). \href{https://youtu.be/NsL4hbSXvFg}{Video link}.
  }
  \label{fig.allegro_results}
\end{figure*}

 
\subsubsection{Task Setup} The Allegro hand on-palm reorientation tasks involve reorienting various objects (Fig. \ref{fig.intro}) from a random  initial pose $(\mathbf{p}^{\text{obj}}_{0}, \mathbf{q}^{\text{obj}}_{0})$ to a randomly given target pose $(\mathbf{p}_{\text{target}}, \mathbf{q}_{\text{target}})$. The  initial pose of an object rests on the hand palm, with the position $\mathbf{p}^{\text{obj}}_{0}$ sampled from the palm center perturbed by a zero-mean Gaussian noise with standard deviation $0.005\boldsymbol{I}$ [m]. The  initial object  orientation  is $\mathbf{q}^{\text{obj}}_{0}{=}\texttt{rpyToQuat}(\phi_0^\text{obj},0, 0)$, with its heading (yaw) angle $\phi_0^\text{obj}$ sampled from
\begin{equation}\label{equ.allegro-init-yaw}
    \phi_0^\text{obj}\sim \mathbb{U}[-\frac{\pi}{2}, \frac{\pi}{2}] \,\, 
    \text{[rad]}.
\end{equation}
To prevent the object from falling, we set the target object position at the center of the palm.
 The  target  orientation is $\mathbf{q}_{\text{target}}{=}\texttt{rpyToQuat}(\phi_\text{target},0, 0)$, with the heading (yaw) angle $\phi_\text{target}$ sampled via 
\begin{equation}\label{equ.allegro-target-yaw}
    \phi_\text{target}\sim \phi_0^\text{obj} + \mathbb{U}(\{-\frac{\pi}{2}, \frac{\pi}{2}\}) \,\, 
    \text{[rad]}.
\end{equation}
That is,  the target orientation a  rotation of the initial orientation $\phi_0^\text{obj}$ by ${\pi}/{2}$ or $-{\pi}/{2}$ uniformly. We choose this large discrete target orientations instead of a continuous range to make the reorientation tasks more challenging for the Allegro hand.

The complementarity-free model setting is  in Table \ref{table.dexhand_model_params}. 
In MPC, the path and final cost functions are  defined as
\begin{equation}\label{equ.allegro_cost_fns}
    \begin{aligned}
        c(\boldsymbol{q}, \boldsymbol{u})&{:=}
        c_\text{contact}(\boldsymbol{q})+ 0.1\norm{\boldsymbol{u}}^2
        \\
        V(\boldsymbol{q})&{:=}
        1000\norm{\mathbf{p}^{\text{obj}}{-}\mathbf{p}_{\text{target}}}^2+50\left(1{-}
        (\mathbf{q}_{\text{target}}\tran\mathbf{q}^{\text{obj}})^2
        \right)
    \end{aligned},
\end{equation}
respectively. 
Here,  $c_{\text{contact}}$ is defined as the sum of the squared distance between object position and all fingertips:
\begin{equation}
 c_\text{contact}(\boldsymbol{q}):=\sum\nolimits_{i=1}^4\norm{
    \mathbf{p}^{\text{obj}}-\mathbb{FK}(\boldsymbol{q}^{\text{finger}_i})
    }^2,
\end{equation}
with $\mathbb{FK}(\boldsymbol{q}^{\text{finger}_i})$ the forward kinematics of finger $i$. Compared to  (\ref{equ.trifinger_cost_fns}), we remove the grasp cost term in $c(\boldsymbol{q},\boldsymbol{u})$ because we found including it is unnecessary for the Allegro hand  tasks here. We also decrease the weight for the position cost in $ V(\boldsymbol{q})$  to prioritize the re-orientation accuracy instead of position. The control bounds  for the Allegro hand are $\mathbf{u}_{\text{lb}}{=-}0.2$ and $\mathbf{u}_{\text{ub}}{=}0.2$. MPC prediction horizon is $T{=}4$.

\begin{figure*}[t]
  \centering
  \includegraphics[width=1\linewidth]{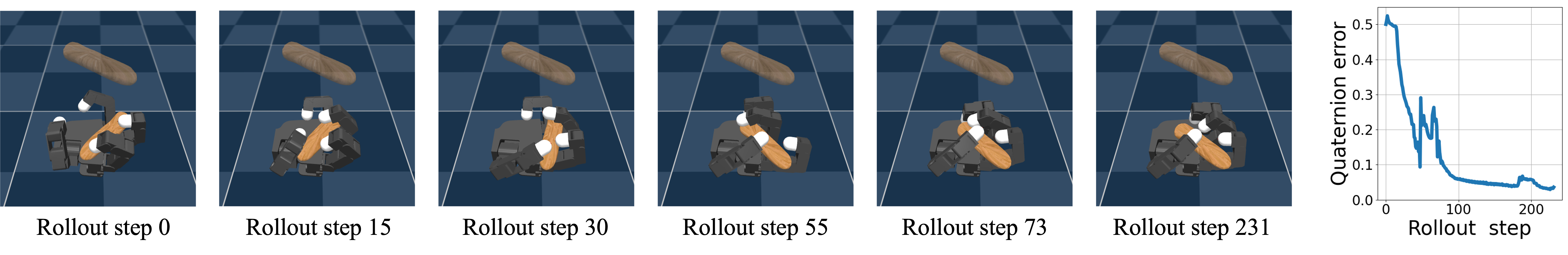}
  \caption{\small An example of the Allegro hand reorientation of a stick object. The first six images display the screenshots of a MPC rollout at different steps, with the target orientation shown in transparency. The last column shows the quaternion error (\ref{equ.fingertips_onground_orierror}) along  rollout steps.
  }
  \label{fig.allegro_example_success}
\end{figure*}

\subsubsection{Results} For each object (Fig. \ref{fig.intro}), we run the MPC policy for 20  trials, each with random initial  and  target poses. The MPC rollout length is set to $H=500$, and the rollout is  terminated early if the following condition,
\begin{equation}\label{equ.allegro_termination}
 \quad1{-}
        (\mathbf{q}_{\text{target}}\tran\mathbf{q}^{\text{obj}})^2\leq 0.04,
\end{equation}
is met consecutively for  20 rollout steps.
We define \begin{equation}\label{equ.allegro_criterion}
    \text{a trial}  
    \begin{cases}
        \text{succeeds} & \text{if (\ref{equ.allegro_termination}) is met for consecutive 20}\\[-2pt]
        & \text{rollout  steps within $H=500$},\\[2pt]
        \text{fails} & \text{otherwise}.
    \end{cases}
\end{equation}

Fig. \ref{fig.allegro_results} presents the Allegro hand on-palm reorientation results for different objects. For each object (x-axis), the upper panel shows the success rate across 20 random trials, while the  bottom panel show the violin plots of final quaternion errors in successful trials. Errors in failed trials are shown as ``gray dots". The final quaternion error is calculated using (\ref{equ.fingertips_onground_orierror}). Table \ref{table.allegro_results_summary} summarizes the overall performance.

\begin{table}[h]
\begin{center}
\begin{threeparttable}
\caption{Summary of the Allegro hand on-palm reorientation results over all objects and all trials}
\label{table.allegro_results_summary}
\begin{tabular}{l l}
     \toprule
     Metric & Value\\
     \midrule
        Overall success rate   & $97.64\% \pm 5.18\%$
          \\
                  Overall final quaternion error    & $0.0116 \pm 0.0088$
                            \\
                  Overall final angle error\tnote{a}    & $11.50^\circ \pm 4.61^\circ$ 
          \\
        Average MPC solving time  per rollout step  & $19 \pm 3$ \text{[ms]}
          \\
     \bottomrule
\end{tabular}
\begin{tablenotes}
\item[a]Angle error is calculated by $\theta{=}\arccos({2(\mathbf{q}_{\text{target}}\tran\boldsymbol{q}^{\text{obj}})^2{-}1})$. 
\end{tablenotes}
\end{threeparttable}
\end{center}
\vspace{-10pt}
\end{table}

\subsubsection{Analysis} Fig. \ref{fig.allegro_results} and Table \ref{table.allegro_results_summary} demonstrate the state-of-the-art performance of the proposed complementarity-free MPC for Allegro hand on-palm reorientation. The overall success rate is $97.64\% \pm 5.18\%$, with an average reorientation error of $11.50^\circ \pm 4.61^\circ$, and MPC runs at over 50 Hz. Fig. \ref{fig.allegro_example_success} shows a  MPC rollout example for   stick reorientation.

\begin{figure}[h]
  \centering
  \includegraphics[width=0.90\linewidth]{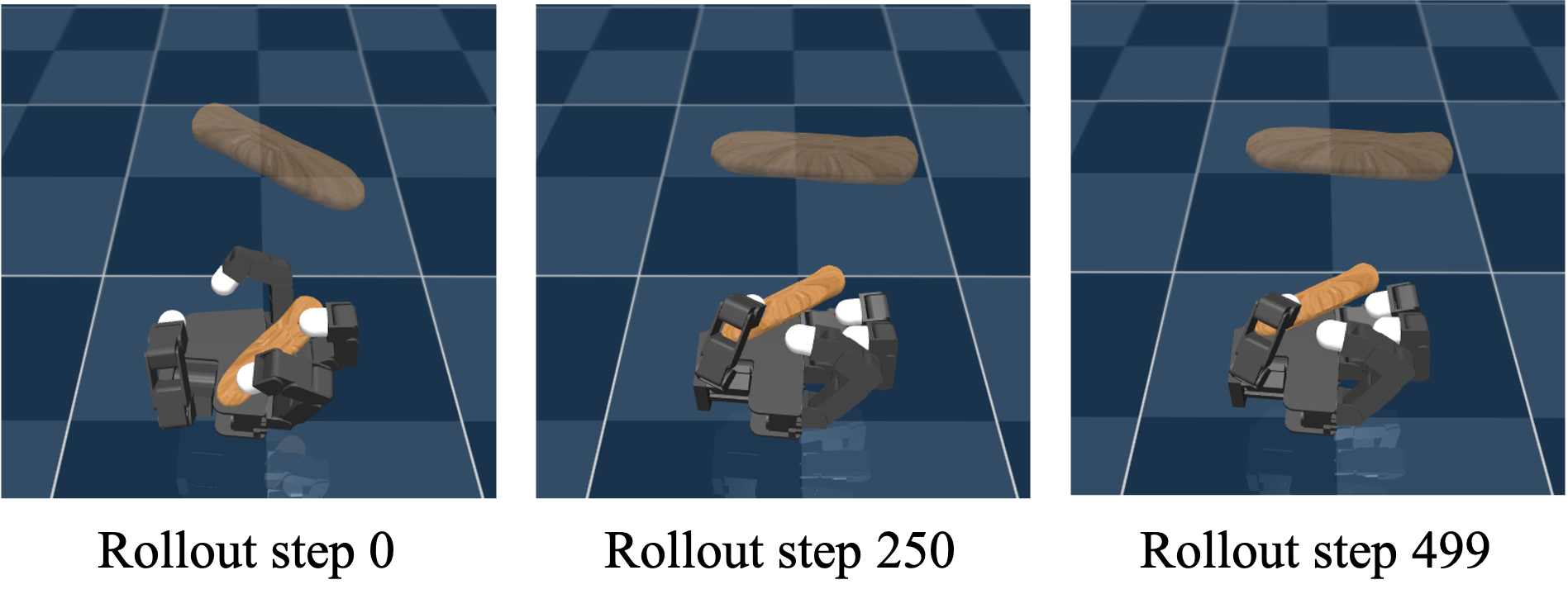}
  \caption{\small An failure case for stick reorientation.}
  \label{fig.allegro_example_failure}
  \vspace{-10pt}
\end{figure}

Fig. \ref{fig.allegro_results} shows lower success rates for the airplane (90\%) and stick (80\%). The primary reason for these failures is the large object geometry relative to the Allegro palm, causing those objects to become out of reach or stuck between the fingers during reorientation. Fig. \ref{fig.allegro_example_failure} gives a failure case.

\subsection{Limitation and Future Work}
In our experiments, we found the contact-implicit MPC is limited to handle dexterous manipulation that requires contact reasoning over \emph{global} geometry. In Algorithm \ref{alg.cf_mpc}, our contact-implicit MPC optimization relies on a fixed and locally linearized geometry (Jacobian). This means the MPC optimizes contact plans using local linearized geometry. Thus, it may struggle with tasks requiring global contact optimization. For example, in peg-in-hole tasks,  successfully  inserting asymmetrical objects into slots requires global contact reasoning.

To address this limitation, one potential direction is to integrate a collision detection pipeline directly into the MPC optimization. However, this approach can pose computational challenges, as collision detection itself can be a non-differentiable optimization problem. Our recent work \cite{yang2024contactsdf}, which introduces an end-to-end smooth contact model encompassing both collision detection and time stepping, may offer a promising solution. Another direction involves incorporating a high-level geometric planner to guide the contact-implicit MPC. Among these directions, the proposed complementarity-free multi-contact model remains a strong candidate for efficient and scalable local contact reasoning. We plan to explore these research directions in future work.

\section{Conclusion}

We propose a complementarity-free multi-contact model for planning and control in dexterous manipulation. By reformulating complementarity constructs into closed forms within the duality of optimization-based contact formulations, our model offers significant computational benefits in contact-rich modeling and optimization: time-stepping explicitness, automatic sanctification with Coulomb friction law, differentiability,  and fewer hyperparameters. The effectiveness of the proposed complementarity-free model and its MPC have been thoroughly evaluated in a range of challenging dexterous manipulation tasks, including fingertips in-air 3D manipulation, TriFinger in-hand manipulation, and Allegro hand on-palm reorientation, all involving diverse objects. The results show that our method consistently achieves state-of-the-art results for highly versatile, high-accuracy, and real-time contact-rich dexterity.

\appendix
\subsection{Proof of Lemma \ref{lemma.dual}}
The KKT optimality condition of the optimization (\ref{equ.dual_prob_relax}) writes
\begin{equation}
    \begin{aligned}\small
        \frac{1}{h^2} \left(
    \boldsymbol{\Tilde{J}} \boldsymbol{Q}^{-1} \boldsymbol{\Tilde{J}}\tran{+}\boldsymbol{R}
    \right)\boldsymbol{\beta} 
    +\frac{1}{h}\left(\boldsymbol{\Tilde{J}} \boldsymbol{Q}^{{-}1}\boldsymbol{b}+{\boldsymbol{\Tilde{\phi}}}\right)-\boldsymbol{\mu}&=\boldsymbol{0},\\
    \boldsymbol{0}\leq \boldsymbol{\mu}\perp\boldsymbol{\beta}&\geq \boldsymbol{0},
    \end{aligned}
\end{equation}
with $\boldsymbol{\mu}$  the vector of the Lagrangian multipliers  for the  constraints in (\ref{equ.dual_prob_relax}). Writing $\boldsymbol{\mu}$ in terms of $\boldsymbol{\beta}$ from the first equation and substituting it the second equation  leads to (\ref{lemma.dual}), which completes the proof. 

\subsection{Proof of Lemma \ref{lemma.key}}
With a positive definite diagonal matrix $\boldsymbol{K}(\boldsymbol{q})$  replacing $(
    \boldsymbol{\Tilde{J}} \boldsymbol{Q}^{-1} \boldsymbol{\Tilde{J}}\tran{+}\boldsymbol{R}
    )^{-1}$  in (\ref{equ.dual_prob_relax_sol}), the approximation  dual solution $\boldsymbol{\beta}^+$ to (\ref{equ.dual_prob_relax_sol}) can be obtained based on the following identity
    \begin{equation}
        0\leq x\perp x+\lambda\geq0 \Leftrightarrow x=\max(-\lambda,0),
    \end{equation}
 which leads to (\ref{equ.key_dual}).
With the approximation dual solution  (\ref{equ.key_dual}), the approximation primal solution $\boldsymbol{v}^+$ in (\ref{equ.key_primal})  can be obtained by substituting $\boldsymbol{\beta}^+$ to (\ref{equ.primal_dual_sol}). This completes the proof.

\bibliographystyle{unsrt}
\bibliography{mybib}

\end{document}